\pdfoutput=1

\documentclass[10pt,twocolumn,letterpaper]{article}

\usepackage[pagenumbers]{cvpr} 
\def\paperID{11200} 
\def\confName{CVPR}
\def\confYear{2025}

\title{Enhancing Creative Generation on Stable Diffusion-based Models}

\author{
Jiyeon Han\thanks{Equally contributed.} $\;^{1}$,\;  Dahee Kwon\footnotemark[1]  $\;^{1}$,\; Gayoung Lee $^{2}$,\; Junho Kim\thanks{Corresponding authors.} $\;^{2}$,\; and Jaesik Choi\footnotemark[2] $\;^{1, 3}$\\
$^1$KAIST AI \quad $^2$ NAVER AI Lab \quad $^3$INEEJI\\
{\tt\small \{j.han, daheekwon, jaesik.choi\}@kaist.ac.kr, \{gayoung.lee, jhkim.ai\}@navercorp.com}
}

\definecolor{cvprblue}{rgb}{0.21,0.49,0.74}
\usepackage[pagebackref,breaklinks,colorlinks,allcolors=cvprblue]{hyperref}

\usepackage{kotex}
\usepackage{multirow}
\usepackage{makecell}
\usepackage{tikz}
\usetikzlibrary{shapes}
\usepackage{caption}
\usepackage{subcaption}
\usepackage{lipsum}
\usepackage[accsupp]{axessibility}
\usepackage[normalem]{ulem}  

\newcommand{\mytriangle}[1]{\tikz{\node[draw=#1,fill=#1,isosceles
triangle,isosceles triangle stretches,shape border rotate=90,minimum
width=0.2cm,minimum height=0.2cm,inner sep=0pt] at (0,0) {};}}

\newcommand{\mydowntriangle}[1]{\tikz{\node[draw=#1,fill=#1,isosceles
triangle,isosceles triangle stretches,shape border rotate=270,minimum
width=0.2cm,minimum height=0.2cm,inner sep=0pt] at (0,0) {};}}

\begin{document}

\makeatletter
\let\@oldmaketitle\@maketitle
\renewcommand{\@maketitle}{\@oldmaketitle
\centering
  \includegraphics[width=0.85\textwidth]
    {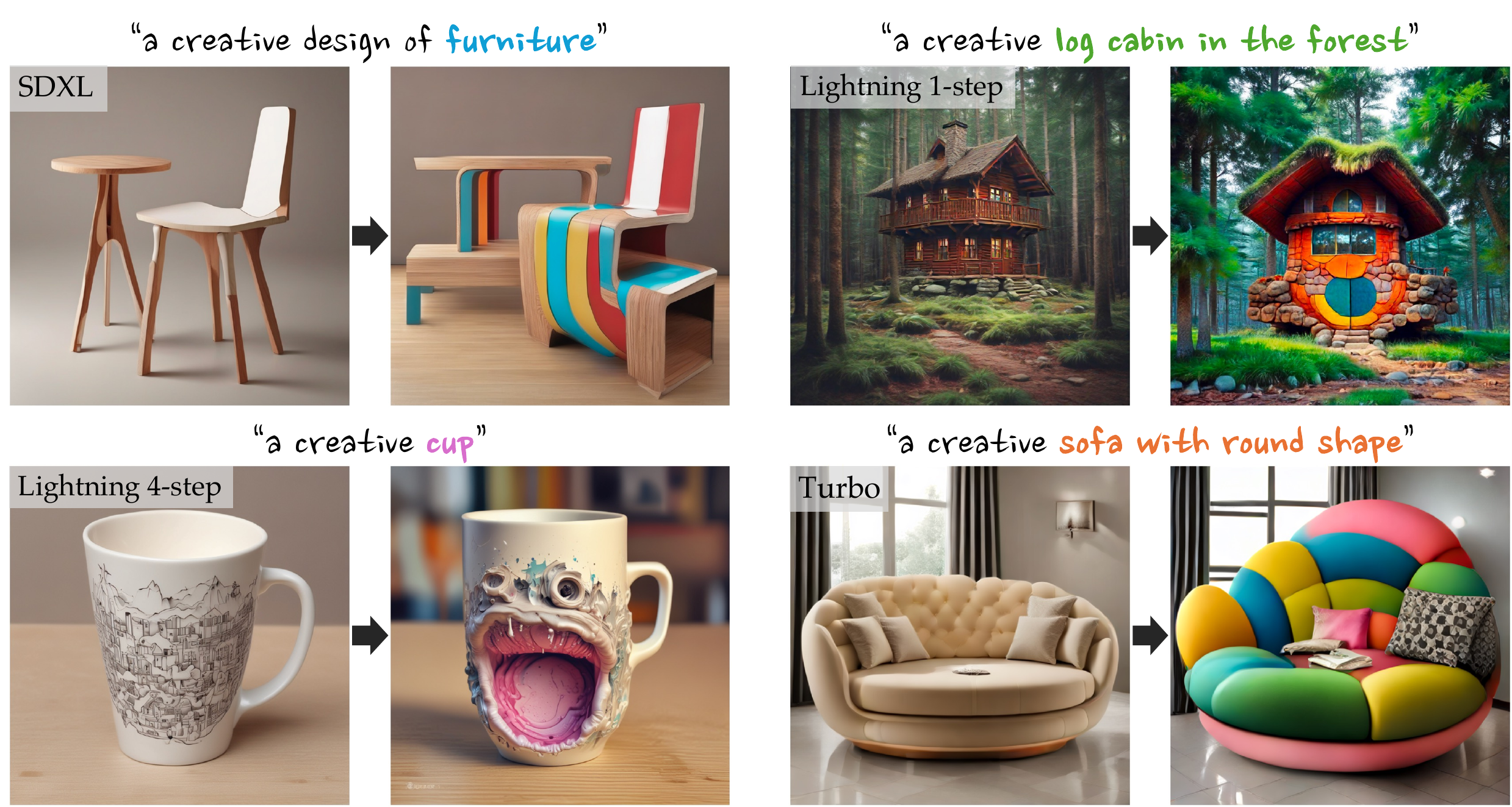}
     \captionof{figure}{\textbf{Original} vs \textbf{C3 (Ours)}. Compared to the original diffusion models, Our C3 consistently generates more creative images with no added computational cost. Code is available at \url{https://github.com/daheekwon/C3}.} 
         \label{fig:page_one}
         \bigskip}
\makeatother
\maketitle

\begin{abstract}
Recent text-to-image generative models, particularly Stable Diffusion and its distilled variants, have achieved impressive fidelity and strong text-image alignment. However, their creative capability remains constrained, as including `creative' in prompts seldom yields the desired results. 
This paper introduces C3 (Creative Concept Catalyst), a training-free approach designed to enhance creativity in Stable Diffusion-based models. C3 selectively amplifies features during the denoising process to foster more creative outputs. We offer practical guidelines for choosing amplification factors based on two main aspects of creativity. 
C3 is the first study to enhance creativity in diffusion models without extensive computational costs. 
We demonstrate its effectiveness across various Stable Diffusion-based models.
\end{abstract}
\vspace{-1.2em}

\section{Introduction}
\label{sec:intro}
\begin{figure*}[!ht]
  \centering
         \includegraphics[width=0.9\textwidth]{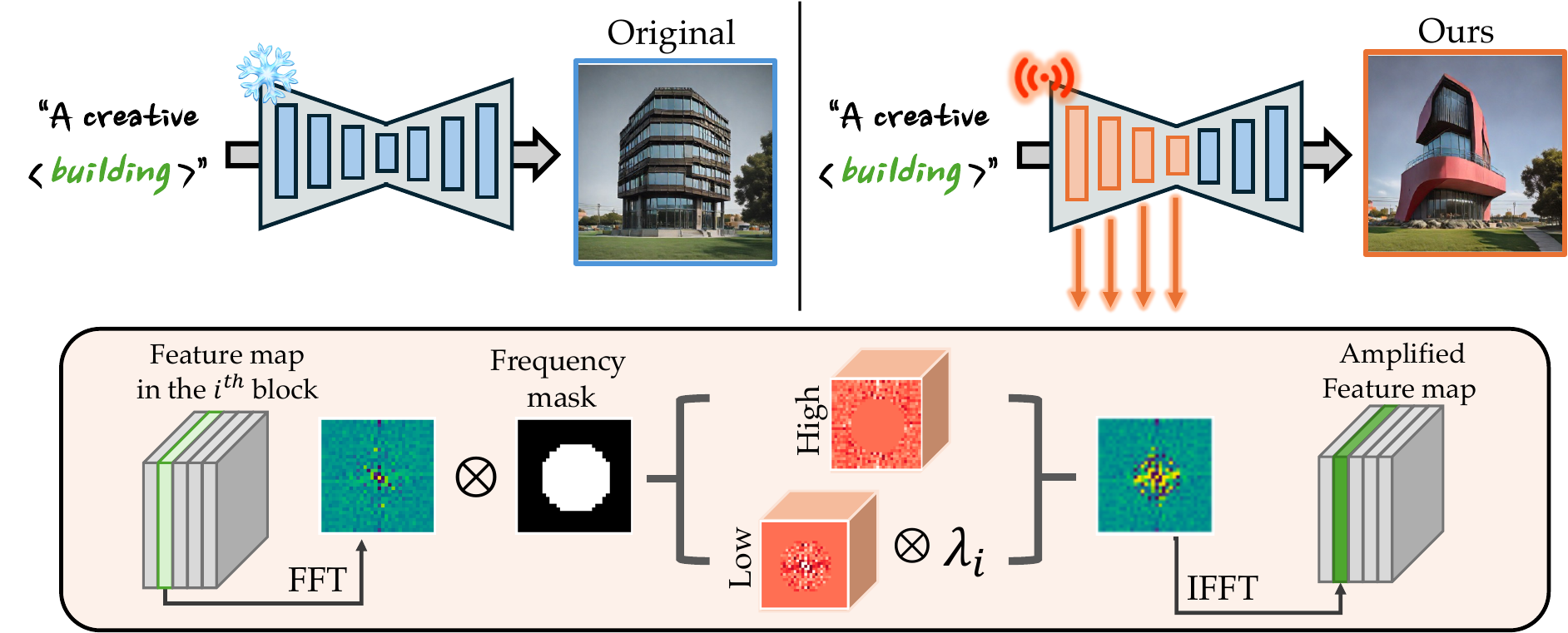}
         \caption{ An overview of the proposed \textbf{C3} algorithm. We selectively amplify the low-frequency feature of the shallow blocks to enhance creative generations of the pretrained diffusion models. 
         }
         \label{fig:main}
\end{figure*}
In recent years, the field of text-to-image generation has witnessed remarkable progress, marked by the development of models capable of producing high-fidelity images that align closely with user-specified text prompts, enabling applications across various fields, including art, entertainment, design, and research~\cite{yang2024diffusionmodelscomprehensivesurvey,wang2024diffusionbasedvisualartcreation}. 
Notably, Stable Diffusion~\cite{rombach2022high,podell2023sdxl} and its distilled variants~\cite{lin2024sdxl,sauer2025adversarial} have emerged as powerful tools, delivering impressive image quality and semantic consistency. As the capabilities of these models continue to expand, so too does the interest in exploring the boundaries of their generative potential, particularly concerning the generation of creative and novel content. This practical pursuit of creativity in generative models has raised critical questions: to what extent can these models foster creativity, and how might we enhance their creative capabilities in an efficient, user-friendly manner?

Despite their prowess, Stable Diffusion-based generative models struggle to effectively produce creative images. As Figure~\ref{fig:page_one} and our user study in Table~\ref{tab:user_study} illustrate, adding the ``creative" term to prompts often fails to yield satisfying creative variations, indicating limitations in the models' creative flexibility. 
While generating creative outputs is a crucial aspect of generative modeling, it remains relatively underexplored. Existing frameworks that aim to improve the novelty of images from diffusion-based models~\cite{richardson2023conceptlab,vinker2023concept,lu2024procreate} typically rely on additional optimization steps or user-defined reference images, making them computationally expensive for scalable use.

To address this gap, we propose a simple yet effective training-free approach, C3 (Creative Concept Catalyst), specifically designed to elevate the creative capabilities of Stable Diffusion-based models. 
Inspired by pioneering work in feature manipulation~\cite{bau2019gan,hertz2022prompt,kwon2022diffusion,si2024freeu,voynov2023p+}, C3 promotes the generation of creative visual concepts by directly amplifying feature maps within the denoising process. In this way, we can bypass the need for reference images.
As will be analyzed in this paper, we observe that different blocks contribute variably to creativity-related image generation, with shallower blocks playing a more significant role in producing novel visuals. Building on this insight, we amplify features at shallower blocks to promote more creative outputs. However, consistent amplification of intermediate features may cause unwanted noise. To address this, we shift the intermediate features into the Fourier domain and selectively manipulate the low-frequency components, effectively boosting creativity while minimizing noise. Also, we offer practical guidelines for selecting appropriate amplification factors tailored to two core aspects of creativity.

To the best of our knowledge, this paper is the first to propose a method for enhancing the creativity of diffusion models without additional optimizing steps. We empirically validate the efficacy of C3 across diverse objects and Stable Diffusion-based models, demonstrating its impact and utility in creative image generation.
\section{Related Work}
\label{sec:related_work}
Research on achieving creative generations in generative models has been continuously advancing. Based on GANs, creative generations are encouraged by employing contrastive loss or diversity loss from existing categories or samples ~\cite{nobari2021creativegan,sbai2018design,elgammal2017can}. 
Recent advances in generative modeling have aimed to balance creativity with diversity in image generation, focusing on approaches that allow inspiration from existing concepts without direct replication. ProCreate~\cite{lu2024procreate}, an energy-based approach, proposes guiding diffusion model outputs away from reference images in the latent space, thus improving diversity and concept fidelity in few-shot settings. This method prevents training data replication and has enhanced sample creativity across various artistic styles and categories. On the other hand, Inspiration Tree~\cite{vinker2023concept} introduces a structured decomposition of concepts, where a hierarchical tree structure captures different visual aspects of a given concept. 
Adding to this line of creative generative techniques, ConceptLab~\cite{richardson2023conceptlab} leverages a Vision-Language Model (VLM) with diffusion priors to further push the boundaries of novel concept generation within broad categories. By iteratively applying constraints that differentiate generated concepts from existing category members, ConceptLab enhances the creation of unique, never-before-seen concepts, enabling hybridization and exploration within a given category. 
While these approaches represent advancements in generating creatively inspired outputs, they require burdensome additional training or optimization. 
An overview of Stable Diffusion-based models and feature map manipulation research is provided in Appendix~\ref{sec:app_related}. 
\section{Methodology}
\label{sec:algorithm}

\subsection{Motivation}

\begin{figure}[t]
  \centering
  
         \includegraphics[width=\columnwidth]{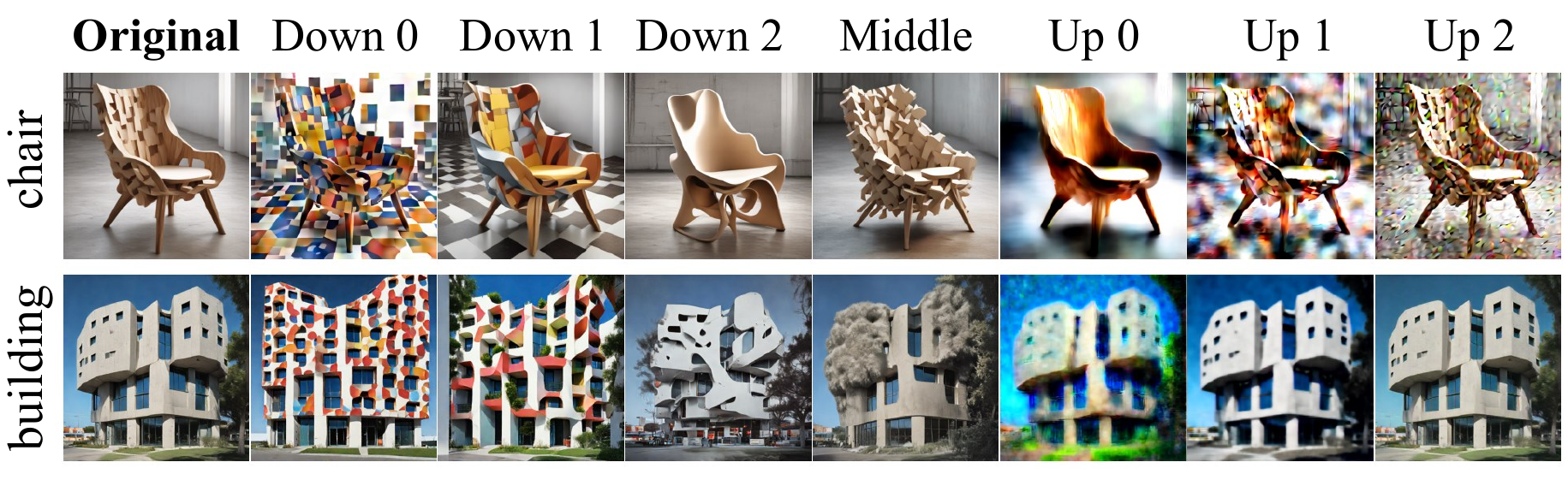}
         \caption{ Block-wise feature amplification results. 
         All frequency bands are amplified. 
         }
         \label{fig:layer_wise}
  
\end{figure}
Stable Diffusion models~\cite{rombach2022high,podell2023sdxl} are widely adopted for their efficient text-to-image generation capabilities, supported by openly accessible checkpoints. Their distilled variants, Turbo~\cite{sauer2025adversarial} and Lightning~\cite{lin2024sdxl} are optimized specifically for faster sampling. 
They share a U-Net backbone with three down blocks, a middle block, and three up blocks to generate latent noise.
For users aiming to create novel and creative images, a straightforward approach is to include the word ``creative" in the prompt. 
However, as shown in Figure~\ref{fig:page_one}, this naive approach proves ineffective across all models. 


Our objective in this paper is to address these limitations and enhance the creative generation capacity of Stable Diffusion-based models. Inspired by feature manipulation methods, we amplify feature maps given a creativity-specified text prompt. If not mentioned otherwise, we use ``a creative [obj]" for the text prompt. As depicted in Figure~\ref{fig:layer_wise}, our empirical analysis demonstrates that each block contributes differently to creativity. Amplifying the first and second blocks primarily induces color and structure changes, while the third down block and middle block impact attributes associated with texture and shape. In contrast, manipulating up blocks mainly affects properties like noise, blur, and contrast. These findings guide our approach: by focusing on the down and middle blocks, we aim to enhance the generation of consistently creative images through targeted feature manipulation.

\subsection{Creative Concept Catalyst (C3)}

\begin{figure}[t]
  \centering
  
         \includegraphics[width=\columnwidth]{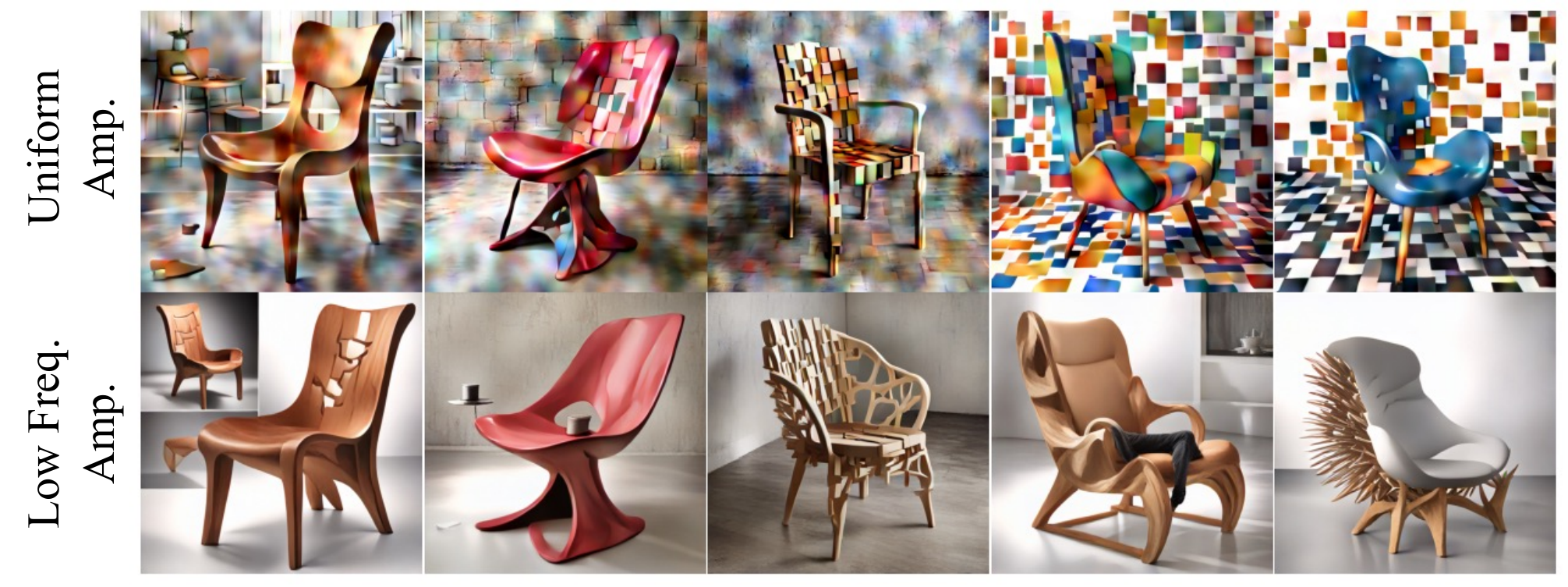}
         \caption{\textbf{(Top)} Uniform amplification across all frequency-band features in the first down block. \textbf{(Bottom)} Amplification of low-frequency features. Enhancing only low-frequency features helps eliminate noise and mosaic patterns.
         }
         \label{fig:frequency}
  
\end{figure}

Here, we introduce a simple yet effective method, C3 (Creative Concept Catalyst), designed to enhance creative image generation without additional training steps. The method overview is depicted in Figure~\ref{fig:main}.


At its core, C3 works by amplifying the internal feature maps in three down blocks and a single middle block within the U-Net. While uniformly amplifying all feature values has the potential to make the image more creative, it often introduces noise and a colorful tile pattern as side effects (See Figure~\ref{fig:frequency}-(Top)). We presume this mainly happens due to the amplification of high-frequency details. In image processing, it is well understood that low-frequency components relate to the main content or objects in an image, while high-frequency components capture finer details. Based on this insight, we selectively amplify the low-frequency components in the frequency domain.

Let $x_l$ denote the output feature maps of the $l$-th block, and $f(x_l) = \text{FFT}(x_l)$ represent the feature maps transformed into the frequency domain. To isolate the low- and high-frequency components, we apply a binarized low-frequency mask $M_L$ with a specified cut-off threshold. The cut-off threshold defines the range of low-frequency components: a higher threshold creates a broader boundary for low frequencies, allowing more extensive modifications in the image, even affecting some finer details. A more detailed analysis of the cut-off threshold, along with our empirical guidelines, is provided in Appendix~\ref{sec:app_hyperparameter}. 
We obtain the low-frequency component $f_L(x_l)$ by multiplying the low-frequency mask with the transformed features $f(x_l)$ element-wise. 
\begin{align}
    f_L(x_l) & = f(x_l) \odot M_L \\
    f_H(x_l) &= f(x_l) \odot (1-M_L)
\end{align}
We amplify the obtained low-frequency components of the feature maps with an amplification factor $\lambda^*_l$ while preserving the high-frequency components. This technique effectively produces a clear but more creatively enhanced object without introducing noise, as illustrated in Figure~\ref{fig:frequency}-(Bottom). The processed features $x^*_l$, transformed back into the spatial domain using the inverse Fourier transform, then serve as the input of the $(l+1)$-th block in the U-Net.

\begin{align}
    f^*(x_l) & = \lambda^*_l \cdot f_L(x_l)  + f_H(x_l) \\
    x_l^* &= \text{IFFT}(f^*(x_l))
\end{align}

\begin{figure}[th]
  \centering
  
         \includegraphics[width=0.9\columnwidth]{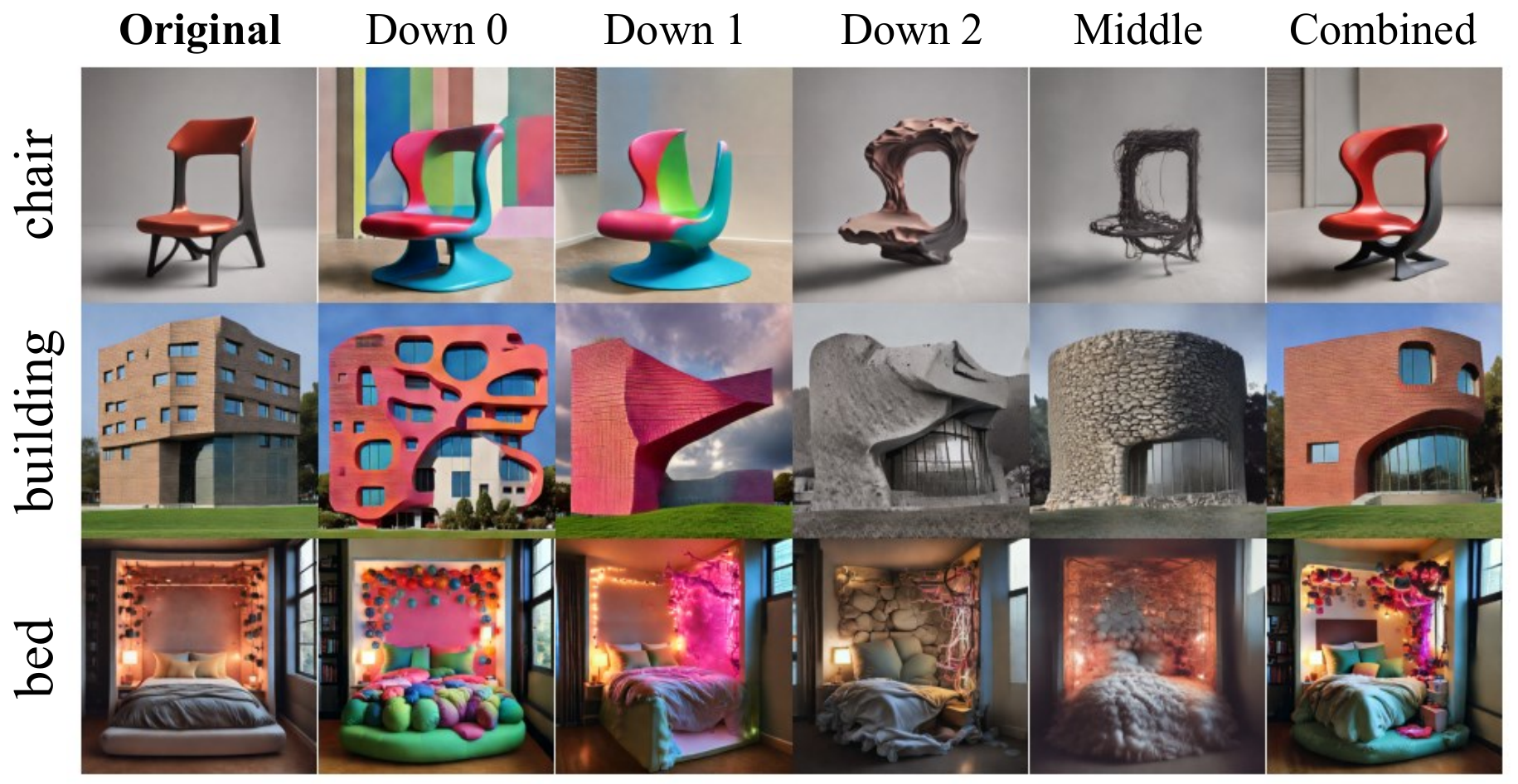}
         \caption{ The image generated with the automatically selected amplification factors for each block and the combined amplification of all blocks.
         }
         \label{fig:combination}

\end{figure}

\subsection{Parameter Selection}
\label{sec:alg_params}
When the amplification factor $\lambda_l$ is too small, the image does not change, and when $\lambda_l$ is too large, the image diverges to an unrecognizable noise. 
Further, the proper amplification factor $\lambda_l$ varies for each block as the later blocks are less sensitive to the change. Here, we propose an automatic way to find a proper amplification factor for each block based on the core aspects of creativity. 

Creativity is known to be recognized in two main aspects: usability and novelty~\cite{mateja2021towards,barron1955disposition}. A creative sample should not only be novel within the population but also meet a certain standard of quality. In the image domain, usability may consider both whether the content of the image can be recognized as the target object and whether the image quality is satisfactory. For instance, while the images in the last column of Figure \ref{fig:layer_wise} can be identified as the target objects, the noise in the image diminishes their usability.
An aesthetic score evaluates image quality based on human perception, potentially considering factors such as color harmony, global layout, and the rule of thirds. 
On the other hand, the CLIP score~\cite{hessel2021clipscore} calculates the similarity between the image embedding and the text embedding in the same space, evaluating how well the image aligns with the given text prompt.
We then define the usability score of an image as follows, 
\begin{equation}
    \text{Use}(I) = Aesthetic(I) + CLIP(I, c)     
\end{equation}
for a generated image $I$ and the text prompt $c$. 
We use a publicly available aesthetic score predictor\footnote{\url{https://github.com/discus0434/aesthetic-predictor-v2-5}} trained on a large-scale database for aesthetic visual analysis (AVA)~\cite{murray2012ava}.  

Assessing the novelty of an image poses more significant challenges, as it requires consideration of all potential outcomes within the population. Rather than measuring novelty directly, we rely on the intuition that novelty increases monotonically with the amplification factor. We then search for the maximum $\lambda_l$ under the usability constraint such that it maximizes novelty while maintaining an acceptable level of quality.
\begin{gather}\label{eq:usability}
    \lambda_l^{*} = \text{max } \lambda_l^i \quad \text{for } \lambda_l^i \in \Lambda_l, \\ 
    \text{s.t.,  Use} (I(\lambda_l^i)) \ge \epsilon\cdot \text{Use}(I(\lambda_l^0)). \nonumber
\end{gather}

Here, $I(\lambda)$ represents the image generated with $\lambda$ amount amplified feature and $\Lambda_l = \{\lambda_l^i | 1=\lambda_l^0 < \lambda_l^i<\lambda_l^{i+1}<...<\lambda_l^n=K_l\}$, where $\lambda_l^0$ represents no amplification and $K_l$ denotes the maximum amplification. 
The usability bumper $0\le\epsilon\le1$ controls the trade-off between usability and novelty—the larger $\epsilon$ results in higher fidelity at the expense of novelty and vice versa. 
For the sake of computational efficiency, we empirically assigned different values for $K_l$ to each block. Typically, we set $K_0=K_1=2$ and $K_2=K_3=10$. Amplification factors larger than these values likely result in noisy images. Nevertheless, it is also feasible to set $K_0=K_1=...=K_L$.  
When amplifying multiple blocks simultaneously, the amplification factors need additional scaling to maintain the image quality. Our empirical findings indicate that the quality is maintained as long as the sum of the scaling factors is preserved. By default, maintaining the sum to 1 for Turbo and 0.6 for the rest of the models generally yields satisfactory results. The detailed analysis of the scaling factors is in Appendix~\ref{sec:app_hyperparameter}. 
The resulting images with feature amplification, using automatically determined amplification factors for each block, along with images amplified across all down and middle blocks, are presented in Figure \ref{fig:combination}.

\begin{table*}[!ht]
  \centering
  \begin{tabular}{@{}c|c|cc|ccc|cc@{}}
    \toprule
      \multirow{2}{*}{Model} &  \multirow{2}{*}{Method} & \multicolumn{2}{c|}{Novelty}    & \multicolumn{3}{c|}{Diversity} & \multicolumn{2}{c}{Usability}    \\ \cline{3-9}
     &   &   FID$^*$ ($\uparrow$) &  Precision$^*$ ($\downarrow$) & Recall ($\uparrow$) & $\text{LPIPS}$ ($\uparrow$) & Vendi ($\uparrow$) & CLIP ($\uparrow$) & BLIP ($\uparrow$)  \\
    \midrule
      \multirow{2}{*}{\makecell{Lightning\\(1-step)}} & Orig & 123.95$\pm$44.94& 0.89$\pm$0.08& 0.69$\pm$0.24& 0.26$\pm$0.10& 5.31$\pm$1.83& \textbf{0.27$\pm$0.02}& \textbf{0.97$\pm$0.04} \\
     & Ours & \textbf{163.01$\pm$58.01}& \textbf{0.65$\pm$0.20}& \textbf{0.79$\pm$0.11}& \textbf{0.34$\pm$0.07}& \textbf{6.24$\pm$2.15}& 0.27$\pm$0.01& 0.89$\pm$0.06 \\\cline{1-9}
     \multirow{2}{*}{Turbo} & Orig & 146.43$\pm$54.48& 0.87$\pm$0.06& 0.27$\pm$0.17& 0.22$\pm$0.06& 3.54$\pm$1.31& \textbf{0.27$\pm$0.02}& \textbf{1.00$\pm$0.00} \\
     & Ours & \textbf{164.07$\pm$55.47}& \textbf{0.51$\pm$0.21}& \textbf{0.68$\pm$0.18}& \textbf{0.36$\pm$0.09}& \textbf{4.93$\pm$1.76}& 0.27$\pm$0.02& 0.95$\pm$0.06 \\\cline{1-9}
     \multirow{2}{*}{\makecell{Lightning\\(4-step)}} & Orig & 117.32$\pm$39.24& 0.86$\pm$0.05& \textbf{0.92$\pm$0.05}& 0.28$\pm$0.08& 5.02$\pm$2.01& \textbf{0.27$\pm$0.02}& \textbf{0.99$\pm$0.01} \\
     & Ours & \textbf{154.91$\pm$49.36}& \textbf{0.55$\pm$0.13}& 0.83$\pm$0.09& \textbf{0.35$\pm$0.05}& \textbf{6.28$\pm$2.18}& 0.26$\pm$0.02& 0.92$\pm$0.06 \\\cline{1-9}
     \multirow{2}{*}{SDXL} & Orig & 128.20$\pm$41.90& 0.79$\pm$0.13& \textbf{0.96$\pm$0.01}& 0.24$\pm$0.05& 6.52$\pm$2.01& \textbf{0.27$\pm$0.02}& \textbf{0.95$\pm$0.04} \\
     & Ours & \textbf{157.93$\pm$38.37}& \textbf{0.66$\pm$0.16}& 0.93$\pm$0.04& \textbf{0.32$\pm$0.04}& \textbf{7.32$\pm$1.91}& 0.27$\pm$0.02& 0.86$\pm$0.07 \\\cline{1-9}
     Real-to-Ref  & - & 248.13$\pm$35.28& 0.56$\pm$0.31& 0.56$\pm$0.21& -& -& -& - \\\cline{1-9}
     ConceptLab & - & 251.27$\pm$61.29& 0.65$\pm$0.20& 0.64$\pm$0.18& 0.37$\pm$0.02& 8.41$\pm$1.26& 0.25$\pm$0.02& 0.32$\pm$0.30 \\ 
\bottomrule
  \end{tabular}
  \caption{Quantitative results averaged over five objects. The Real-to-Ref row employs the prompt ``a [ref-obj]" to generate reference fake samples. FID$^*$ and Precision$^*$ scores are interpreted in opposition to their conventional usage, as our method aims to generate novel samples distinct from the normal ones. \textbf{Bold} indicates the best for each metric.}
  \label{tab:quantitative}
\end{table*}

\section{Experimental Results}
\label{sec:experiment}

\subsection{Settings}
This section evaluates whether the proposed C3 method can generate sufficiently creative images across various Stable Diffusion-based models and object types. We selected four base models for this analysis: SDXL~\cite{podell2023sdxl} and its distilled versions, Lightning~\cite{lin2024sdxl} (1-step and 4-step) and Turbo~\cite{sauer2025adversarial}.
We also included ConceptLab~\cite{richardson2023conceptlab} as another baseline. Like our approach, ConceptLab generates creative images without requiring reference images. It uses Kandinsky~\cite{razzhigaev2023kandinsky} as the backbone architecture and incurs additional training costs. Detailed descriptions of the various hyperparameter settings are provided in Appendix~\ref{sec:app_settings}. 
Additionally, we provide an analysis of the step-wise effects of C3 at different stages in Appendix~\ref{sec:app_hyperparameter}, while the default setting applies C3 across all denoising steps.

\begin{figure*}[ht]
  \centering
  
         \includegraphics[width=0.95\textwidth]{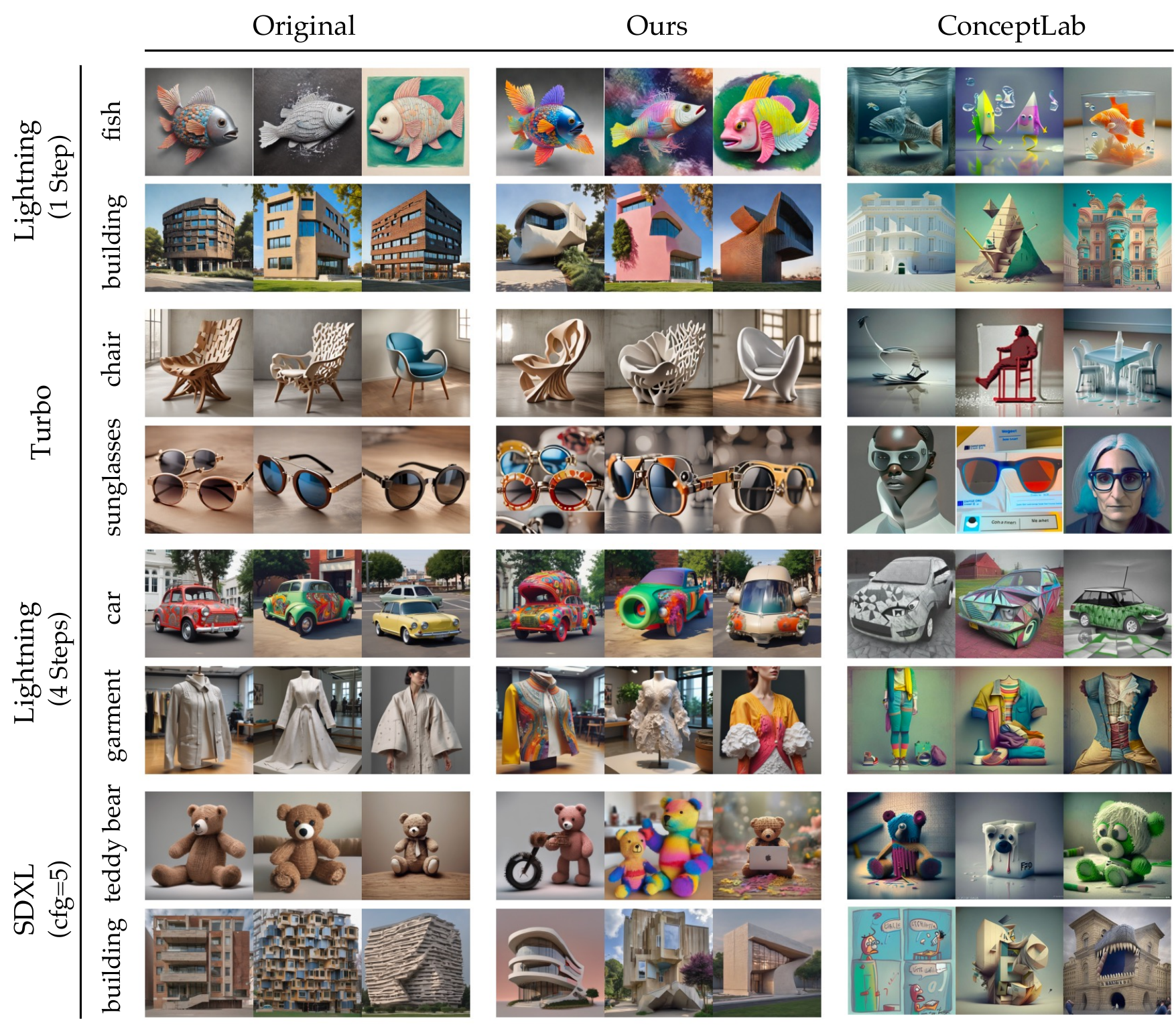}
         \caption{Qualitative results of \textbf{C3}, in comparisons of original generations and ConceptLab. For each object, three images are curated from 100 generations from each model/method. For ConceptLab, we train 10 different individual concepts and generate using 10 random seeds for each concept. 
         }
         \label{fig:qual}
  
\end{figure*}

\subsection{Quantitative Results}
\label{sec:quan}
In this section, we quantitatively analyze the performance of the proposed method across five distinct objects: chair, building, garment, car, and teddy bear. We use 100 uncurated images for each object. 
We categorize the evaluation metrics into three groups: novelty, diversity, and usability. FID computes the distributional similarity between real and fake datasets. Precision and recall measure the proportion of fake data within the real manifold and the proportion of real data within the fake manifold, respectively. In place of the real data, we use the images generated from SDXL with the text prompt ``a [obj]".   

Generally, a lower FID score indicates better performance, and a higher precision is preferred. However, in this study, we aim for outcomes that deviate from normal generations. Consequently, we interpret FID and precision in the opposite manner, where a higher FID is considered better, and a lower precision is favored. 
We establish reference values for FID and precision to prevent misleading results, which are computed on a separate set of samples generated with the prompt ``a [ref-obj]". The reference objects are selected to belong to the same category as the target object while being significantly distinguishable from it. The used reference objects are sofa ($\leftrightarrow$ chair), monument ($\leftrightarrow$ building), scarf ($\leftrightarrow$ garment), bus ($\leftrightarrow$ car), and bunny doll ($\leftrightarrow$ teddy bear). 
Recall indicates how many modes are covered by the generated data, considering each real as each mode. LPIPS measures the perceptual distance between two images using image features extracted from a pre-trained backbone. In our analysis, we measure the average LPIPS between all pairs of generated images. The Vendi score measures the Shannon entropy of the eigenvalues of the similarity matrix between the generated images and can be interpreted as the number of effective modes. For usability, we use the CLIP score between the generated image and the text prompt ``a creative [obj]" used for the generation. The BLIP score measures the portion of generated samples that receive a `yes' response from the BLIP VQA model when asked, ``Is this image [obj]?" 

Table \ref{tab:quantitative} summarizes the results. We average each score across five objects. The scores for each object are listed in Appendix~\ref{sec:app_detailed_exp}.
For the novelty, both FID and Precision values are improved compared to the original outcomes. FID of ConceptLab is larger than ours, however, it exceeds the FID of the reference object, which means the result may be regarded as a different object. 
Regarding diversity, our metrics have shown improvements in general. The recall values for Lightning (4-step) and SDXL have experienced a slight decline but remain at satisfactory levels. Conversely, Turbo and Lightning have significantly enhanced recall scores, particularly the Turbo model, which has been notably affected by mode collapse.
Considering the improvements of our method in terms of novelty and diversity, the loss in usability metrics is tolerable compared to ConceptLab.

\subsection{Qualitative Results}

We demonstrate the capacity of C3 to generate diverse creative concepts across multiple categories. Our results are compared with those of the original Stable Diffusion-based models and ConceptLab~\cite{richardson2023conceptlab}. We generate 100 images for each case, selecting three carefully curated examples for visualization in Figure~\ref{fig:qual}. Uncurated versions can be found in Appendix~\ref{sec:app_detailed_exp}.
Each pair of images from the original models and C3 shares the same random seed, enabling a direct comparison of the changes introduced by the proposed method.

Original models have difficulty producing creative images even with prompts like ``a creative obj". By incorporating our C3 method into these original models, we observe a marked enhancement in their creative generation capabilities. The images generated with our approach preserve the core semantic meaning of the original object while adding richer and more inventive elements. ConceptLab also generates distinct variations but often does so at the expense of the original object’s semantic integrity. These results provide evidence of the considerable advancements achieved with the C3 method, underscoring its effectiveness in enhancing the creative generation capabilities of the model.

\begin{table}[!ht]
  \centering
  \begin{tabular}{@{}c|c|ll@{}}
    \toprule
      Model &  Method & Usability    & Novelty \\ 
      \midrule
       \multirow{2}{*}{Lightning (1-step)} & Orig & \textbf{4.62}& 2.65 \\
     & Ours & 4.19 (\mydowntriangle{blue}0.43)& \textbf{4.12} (\mytriangle{red}1.47) \\
\midrule
     \multirow{2}{*}{Turbo} & Orig & \textbf{4.49}& 3.08 \\
     & Ours & 4.14 (\mydowntriangle{blue}0.35)& \textbf{3.79} (\mytriangle{red}0.71) \\
\midrule
\midrule
     ConceptLab & - & 2.97& 3.65 \\ 
\bottomrule
  \end{tabular}
  \caption{User study results averaged over five objects. `Usability' evaluates whether the image accurately represents the specified [obj], while `Novelty' assesses the image's uniqueness. Responses are collected on a 5-point Likert scale. }
  \label{tab:user_study}
\end{table}

\subsection{User Study}
We conducted a user study to evaluate the creativity of the generated images based on human perception. Participants were asked to respond to two questions for each instance, assessing both usability and novelty.
The user study was conducted for the same five objects detailed in Section \ref{sec:quan}. 
The results are summarized in Table \ref{tab:user_study}. Compared to the baseline models, our approach demonstrates improvements in novelty for both models. Notably, the novelty score surpasses that of ConceptLab. Although there are decreases in usability scores, these losses are smaller than the increases in novelty scores. Furthermore, the usability scores of our method are significantly higher compared to those of ConceptLab. More detailed results and experimental settings can be found in Appendix~\ref{sec:app_detailed_exp}.

\section{Discussions}
\label{sec:discussion}

\subsection{How C3 Enhances Creativity}

\begin{figure}[h!]
  \centering
         \includegraphics[width=\columnwidth]{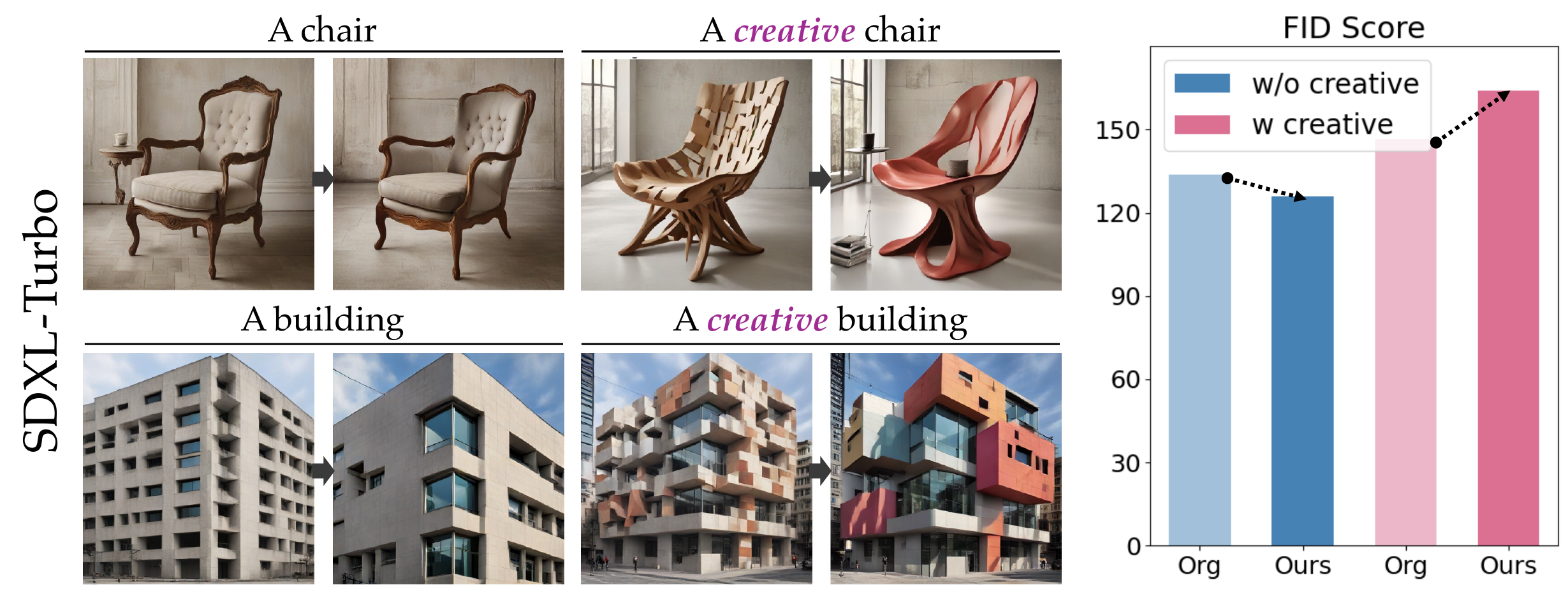}
         \caption{(Left) Images generated with and without ``creative" in prompts.  (Right) Change in FID scores after applying C3.}
         \label{fig:abl_creative}  
\end{figure}

\textbf{``Creative'' Ablation.} We hypothesize that C3 only amplifies creativity when the prompt explicitly includes ``creative.'' Enlarging the latent boosts cross-attention values associated with the creative object, enhancing its generation. To verify this, we ablate the ``creative" in prompt and find that its removal results in ordinary images without noticeable artifacts (see Figure~\ref{fig:abl_creative}). The FID score even decreases, indicating that C3 does not generate unexpected images without ``creative" in prompt. 

\noindent\textbf{Justification for C3.} 
To further support our hypothesis, we examine whether amplifying first-block features increases the second-block cross-attention map of ``[obj]" as ``[obj]" implies ``creative [obj]" due to the self-attention in the text encoder. 
The \textit{t}-test results confirm a significant increase for the cross-attention map of ``[obj]."
in both Turbo and Lightning model with p-value $<e^{-30}$.
C3 amplifies only low-frequency features, which clearly enhance creativity in contents, as creativity driven by high-frequency features appears as colorful mosaics or noise (see Figure 4 in Sec. 3). Furthermore, we automatically select the amplification factor based on the usability score to maintain generation quality.

\subsection{Types of Creativity}
\label{sec:exp_creativity_type}
\begin{figure}[h!]
  \centering
  
         \includegraphics[width=0.9\columnwidth]{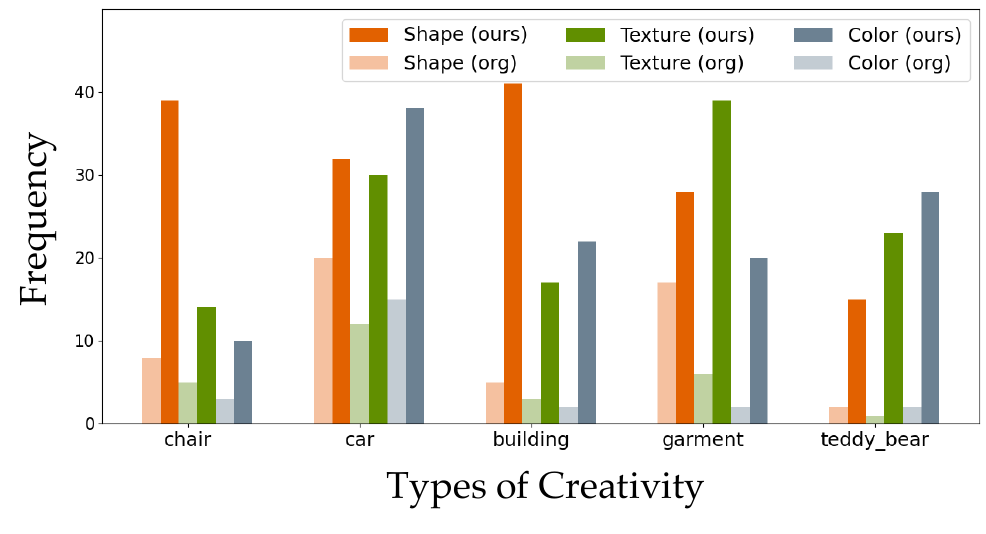}
         \caption{
         Types of creativity classified by GPT 4o for Lightning (1-step) generations. Responses are multiple-choice, among `Shape', `Texture', and `Color'. 50 images are used for each method.
         }
         \label{fig:type}
  
\end{figure}

We investigated which aspects of creativity are enhanced in images generated by the proposed method.
This identification was performed using a multi-modal LLM (GPT 4o~\cite{achiam2023gpt}) to extract responses. The statistical findings, displayed in Figure~\ref{fig:type}, reveal that images generated by our method, depicted in vivid colors, achieve significantly higher values than those from the original model, shown in muted colors, highlighting an overall increase in creativity. Furthermore, we observe that different creative aspects are emphasized for each object type. For example, in the case of a chair, creativity is enhanced primarily in the shape aspect, garments in texture, and teddy bears in color. This observation suggests that our model effectively identifies and emphasizes suitable creative attributes for various objects, enhancing the generated images accordingly.

\subsection{Use Cases}
\begin{figure}[h!]
  \centering
         \includegraphics[width=\columnwidth]{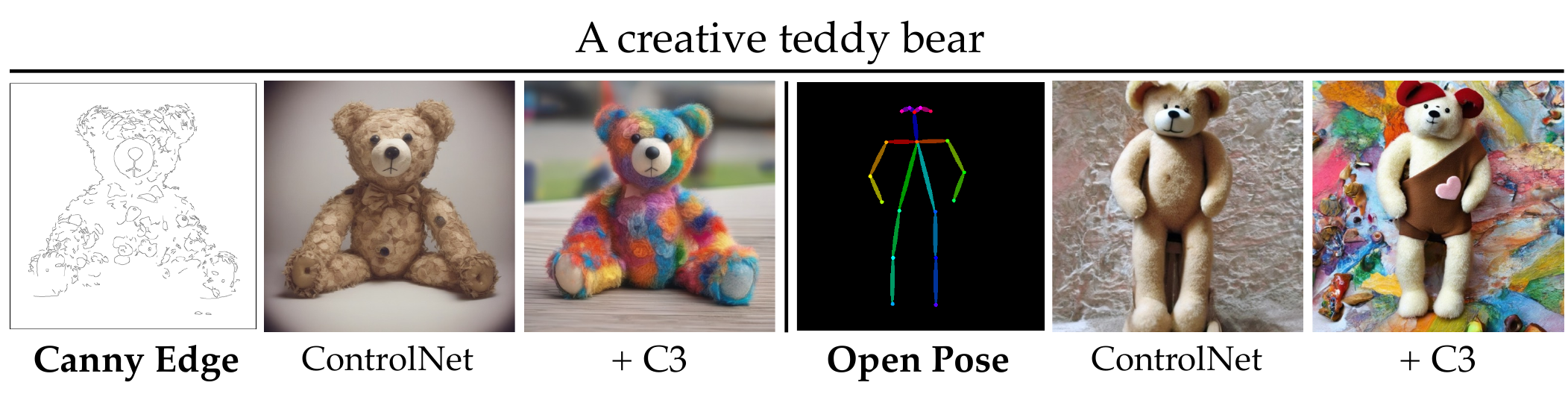}
         \caption{Combining C3 with ControlNet on SDXL.}
         \label{fig:ctrlnet_1}  
\end{figure}

\noindent\textbf{Integration with ControlNet~\cite{zhang2023adding}.} To demonstrate the versatility of C3, we integrate it with ControlNet, a widely recognized plugin adapter. As shown in Figure~\ref{fig:ctrlnet_1}, this combination enables effortless generation of creative samples while adhering to input constraints. 

\begin{figure}[ht]
  \centering
  
         \includegraphics[width=\columnwidth]{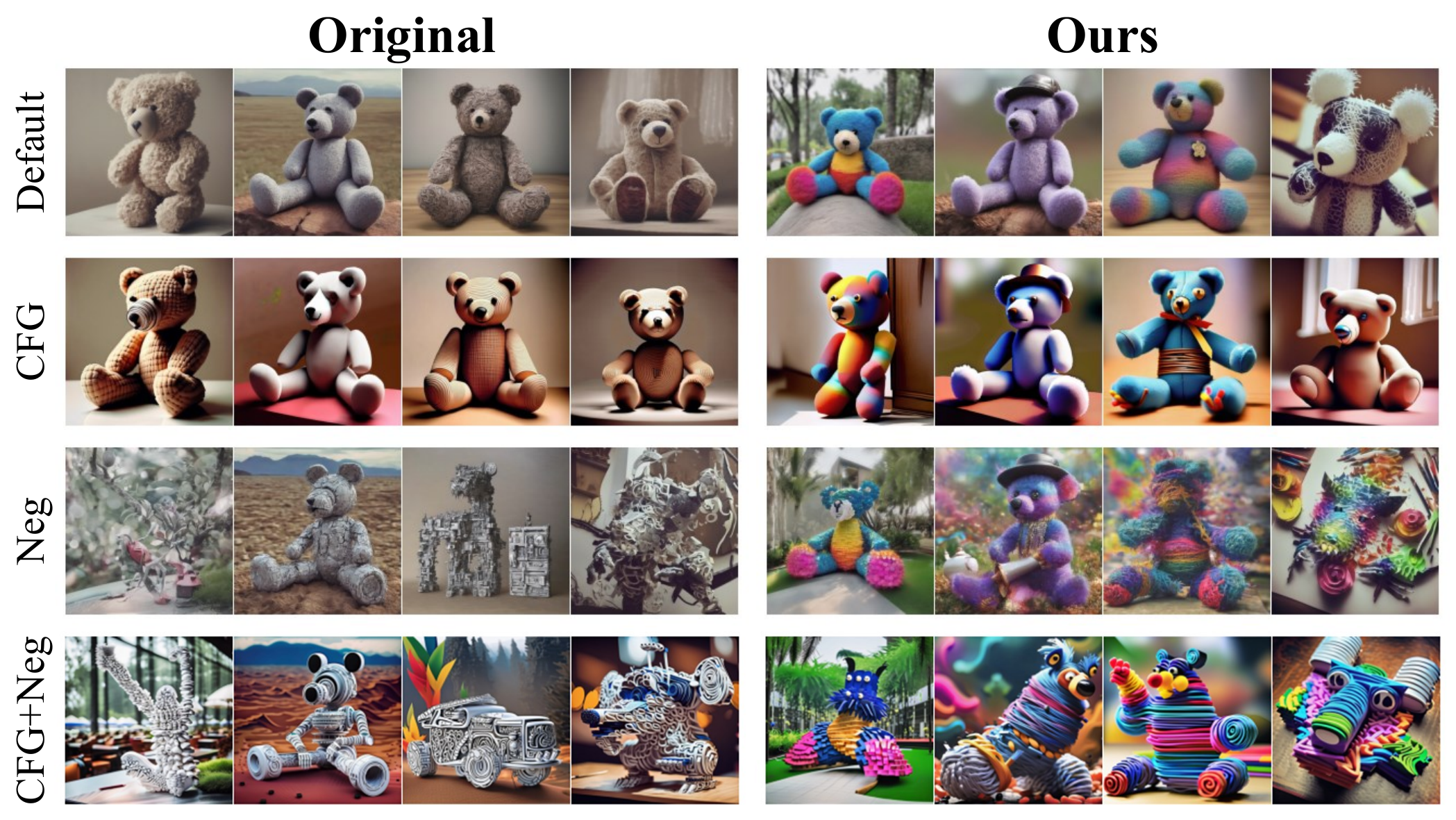}
         \caption{
         Comparisons between pre-defined Hyperparameter controls of SDXL and \textbf{C3}. \textbf{(Default)} CFG=5 / Negative prompt=``". \textbf{(CFG)} CFG=30. \textbf{(Neg)} Negative prompt=``normal teddy bear".
         }
         \label{fig:cfg}
  
\end{figure}

\noindent\textbf{Integration with SDXL Hyperparameters.}
We compare the results of our method with the readily available hyperparameters of the SDXL model, specifically in terms of classifier-free guidance (CFG) scale~\cite{ho2022classifierfreediffusionguidance} and negative prompts. CFG is a hyperparameter that regulates the blend between the outputs of the given prompt and the negative prompt. Higher CFG values are known to enhance text-image alignment. However, we have observed that, in some cases, there are only marginal improvements in creativity, even with large CFG values. As illustrated in Figure \ref{fig:cfg}, increasing the CFG from the default setting of 5 to 30 or using a negative prompt such as ``normal teddy bear" leads to minimal changes in creativity. In contrast, our method enhances the default results without utilizing CFG or negative prompts. Notably, our method can also be employed alongside CFG and negative prompts, yielding more creative samples across all four settings.

\begin{figure}[ht]
  \centering
  
         \includegraphics[width=0.75\columnwidth]{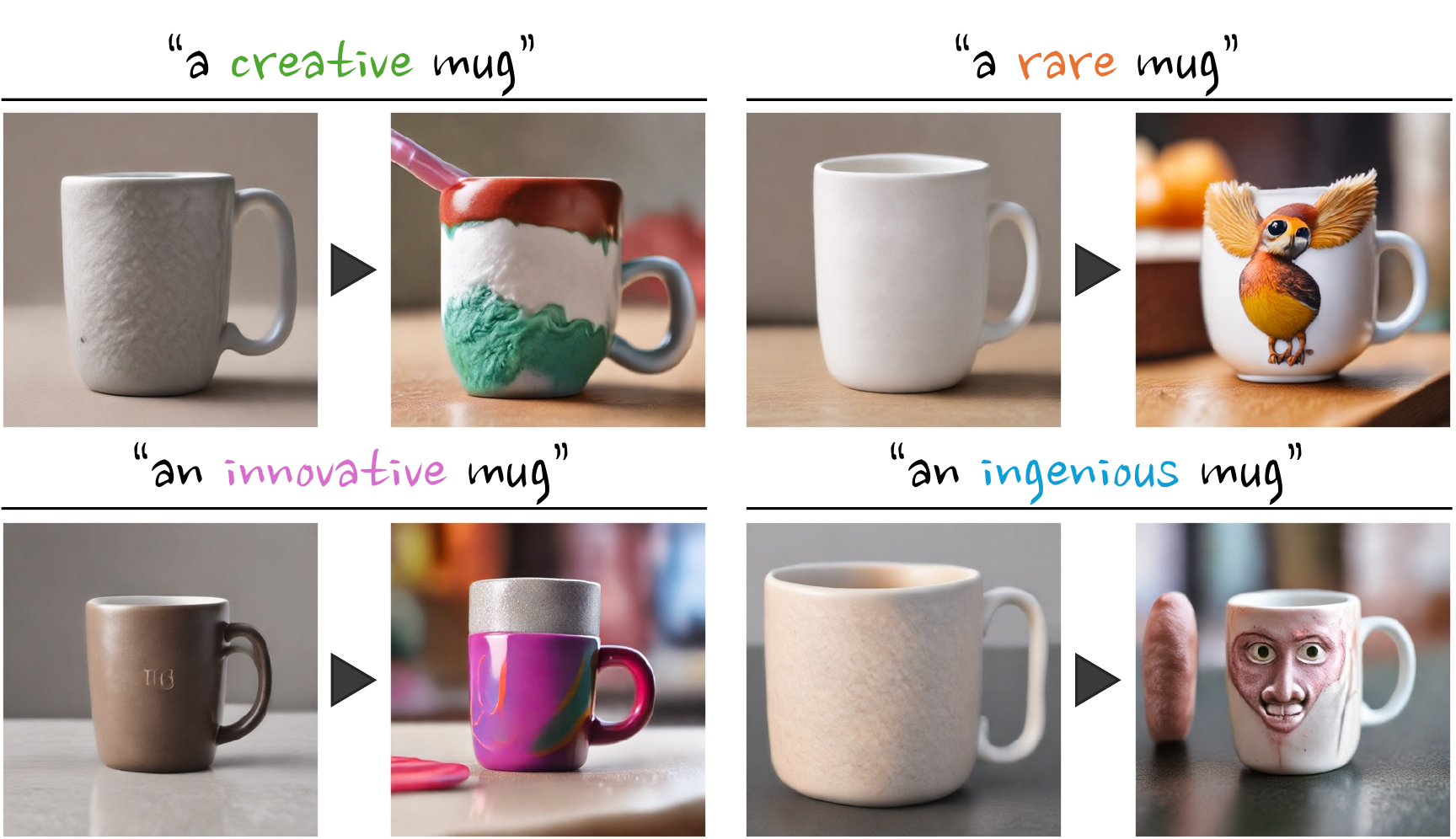}
         \caption{Example outcomes using alternative prompts in place of ``creative." 
         }
         \label{fig:template}
\end{figure}

\noindent \textbf{Extension to Alternative Prompts.} Our method is designed to function with textual descriptions that include \textit{``creative"}. To assess the versatility of our approach, we tested whether similar effects could be achieved using alternative templates beyond ``creative." In Figure~\ref{fig:template}, we constructed prompts using a total of four similar adjectives, $\{\textit{creative, rare, innovative, ingenious}\}$ and analyzed the outcomes when integrated with C3. We observed that across all templates, distinct and creative characteristics were expressed just as effectively as with ``creative." This suggests that C3 consistently enhances creativity across a range of creativity-associated prompts.

\subsection{C3 on Non-Stable-Diffusion Models}
\begin{figure}[h]
  \centering
         \includegraphics[width=0.8\columnwidth]{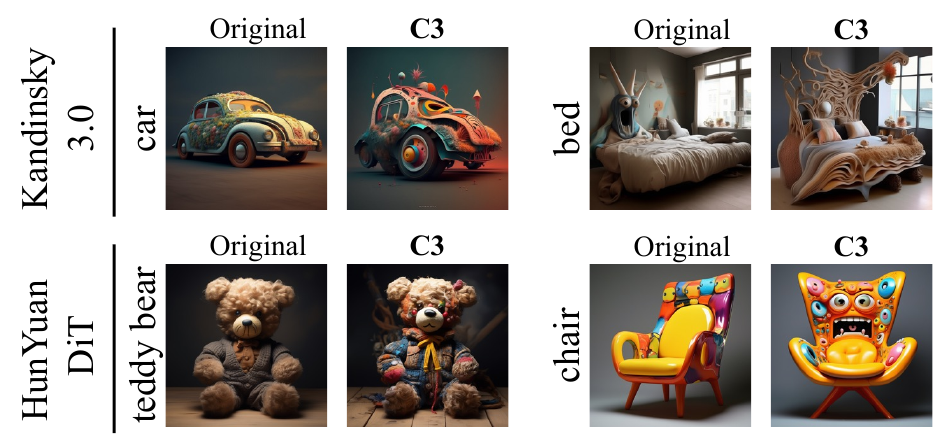}
         \caption{Application of C3 on Kandinsky 3.0 and HunYuan-DiT models. }
         \label{fig:nonsd}  
\end{figure}

 We applied C3 to Kandinsky 3.0 and HunYuan-DiT, in addition to SDXL variants, as presented in Figure~\ref{fig:nonsd}. Kandinsky 3.0 is based on a U-Net structure as SDXL, while HunYuan-DiT is based on a stacked transformer structure. 
 The block-wise analysis for each model is provided in Appendix~\ref{sec:app_nonsd}. 
 While the results open the possibility of expanding C3 to non-SDXL-based models, the specific architecture and components of the model may affect the application of C3 to other models, and more comprehensive analyses might be needed, especially to understand transformer-based models, which we leave as future work.

\vspace{-0.2cm}

\subsection{Comparison with FreeU}

\begin{figure}[h!]
  \centering
         \includegraphics[width=\columnwidth]{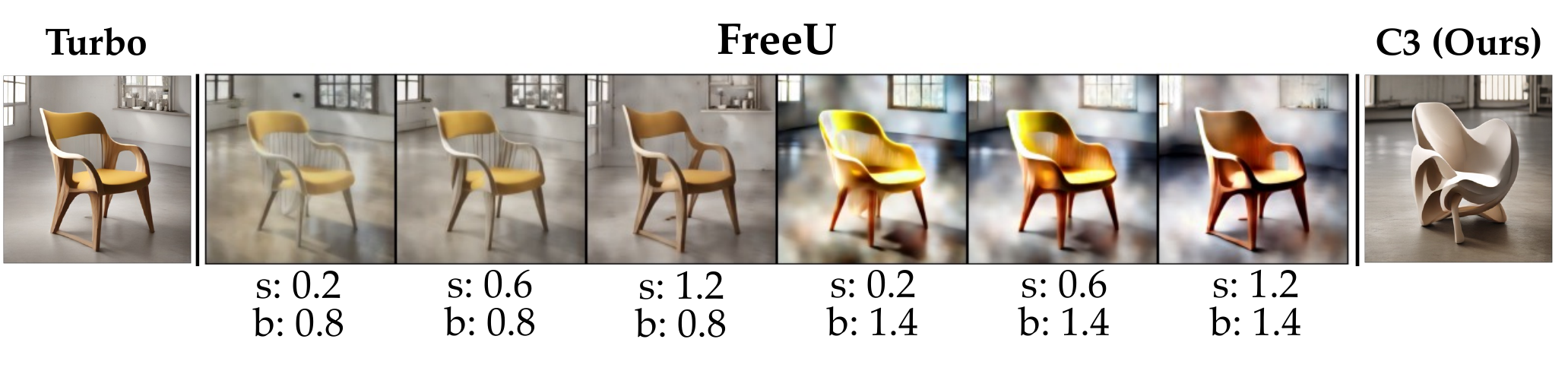}
         \vspace{-0.8cm}
         \caption{Results with prompt  ``A creative chair'' in SDXL-Turbo.}
         \label{fig:freeu}
  
\end{figure}

Unlike FreeU~\cite{si2024freeu}, which enhances image fidelity by modifying low-frequency components in skip connections and boosts backbone features without frequency-based control—using uniform parameters $s$ and $b$ across blocks—C3 specifically targets low-frequency backbone features in the down and middle blocks, applying block-specific parameters. This distinction allows C3 to better steer creativity. Figure~\ref{fig:freeu} shows that FreeU produces less clean and creative results than C3, regardless of parameter values. Especially when $b > 1$, noise persists regardless of $s$.


\section{Conclusion}
\label{sec:conclusion}
In this paper, we propose a simple yet effective method to enhance the creative outputs of pre-trained Stable Diffusion-based models. 
C3 boosts creativity by amplifying internal features with auto-selected amplification factors, preserving quality without extensive fine-tuning or extra optimization.
There are, however, limitations to our method. 
Our method heavily relies on the generation capabilities of the pre-trained models. Acting as a catalyst, our method may fail to generate a creative sample if the model itself has a limited concept of creativity for the target object. 
Moreover, the effectiveness of C3 across different model architectures and components requires comprehensive analysis, which is reserved for future work.
Despite these limitations, we believe that our work, as the first training-free method for enhancing creative generations, can significantly contribute to the creative AI research community and inspire users, such as product designers, through the improved outcomes produced by our method.

\clearpage
\section*{Acknowledgment}
This work was partly supported by the KAIST-NAVER Hypercreative AI Center, the Korean Institute of Information \& Communications Technology Planning \& Evaluation, and the Korean Ministry of Science and ICT under grant agreement No. RS-2019-II190075 (Artificial Intelligence Graduate School Program (KAIST)), No. RS-2022-II220984 (Development of Artificial Intelligence Technology for Personalized Plug-and-Play Explanation and Verification of Explanation), No.RS-2022-II220184 (Development and Study of AI Technologies to Inexpensively Conform to Evolving Policy on Ethics), No. RS-2024-00457882 (AI Research Hub Project) and Center for Applied Research in Artificial Intelligence (CARAI) grant funded by DAPA and ADD (UD230017TD)
{
    \small
    \bibliographystyle{ieeenat_fullname}
    \bibliography{main}

\begin{thebibliography}{31}
\providecommand{\natexlab}[1]{#1}
\providecommand{\url}[1]{\texttt{#1}}
\expandafter\ifx\csname urlstyle\endcsname\relax
  \providecommand{\doi}[1]{doi: #1}\else
  \providecommand{\doi}{doi: \begingroup \urlstyle{rm}\Url}\fi

\bibitem[Achiam et~al.(2023)Achiam, Adler, Agarwal, Ahmad, Akkaya, Aleman, Almeida, Altenschmidt, Altman, Anadkat, et~al.]{achiam2023gpt}
Josh Achiam, Steven Adler, Sandhini Agarwal, Lama Ahmad, Ilge Akkaya, Florencia~Leoni Aleman, Diogo Almeida, Janko Altenschmidt, Sam Altman, Shyamal Anadkat, et~al.
\newblock Gpt-4 technical report.
\newblock \emph{arXiv preprint}, 2023.

\bibitem[Barron(1955)]{barron1955disposition}
Frank Barron.
\newblock The disposition toward originality.
\newblock \emph{The Journal of Abnormal and Social Psychology}, 51\penalty0 (3):\penalty0 478, 1955.

\bibitem[Bau et~al.(2019)Bau, Zhu, Strobelt, Zhou, Tenenbaum, Freeman, and Torralba]{bau2019gan}
David Bau, Jun~Yan Zhu, Hendrik Strobelt, Bolei Zhou, Joshua~B Tenenbaum, William~T Freeman, and Antonio Torralba.
\newblock Gan dissection: Visualizing and understanding generative adversarial networks.
\newblock In \emph{ICLR}, 2019.

\bibitem[Cao et~al.(2023)Cao, Wang, Qi, Shan, Qie, and Zheng]{cao2023masactrl}
Mingdeng Cao, Xintao Wang, Zhongang Qi, Ying Shan, Xiaohu Qie, and Yinqiang Zheng.
\newblock Masactrl: Tuning-free mutual self-attention control for consistent image synthesis and editing.
\newblock In \emph{ICCV}, pages 22560--22570, 2023.

\bibitem[Elgammal et~al.(2017)Elgammal, Liu, Elhoseiny, and Mazzone]{elgammal2017can}
Ahmed Elgammal, Bingchen Liu, Mohamed Elhoseiny, and Marian Mazzone.
\newblock Can: Creative adversarial networks generating “art” by learning about styles and deviating from style norms.
\newblock In \emph{ICCC}, 2017.

\bibitem[Hertz et~al.(2023)Hertz, Mokady, Tenenbaum, Aberman, Pritch, and Cohen-or]{hertz2022prompt}
Amir Hertz, Ron Mokady, Jay Tenenbaum, Kfir Aberman, Yael Pritch, and Daniel Cohen-or.
\newblock Prompt-to-prompt image editing with cross-attention control.
\newblock In \emph{ICLR}, 2023.

\bibitem[Hessel et~al.(2021)Hessel, Holtzman, Forbes, Le~Bras, and Choi]{hessel2021clipscore}
Jack Hessel, Ari Holtzman, Maxwell Forbes, Ronan Le~Bras, and Yejin Choi.
\newblock Clipscore: A reference-free evaluation metric for image captioning.
\newblock In \emph{EMNLP}, pages 7514--7528, 2021.

\bibitem[Ho and Salimans(2022)]{ho2022classifierfreediffusionguidance}
Jonathan Ho and Tim Salimans.
\newblock Classifier-free diffusion guidance, 2022.

\bibitem[Ho et~al.(2020)Ho, Jain, and Abbeel]{ho2020denoising}
Jonathan Ho, Ajay Jain, and Pieter Abbeel.
\newblock Denoising diffusion probabilistic models.
\newblock \emph{NeurIPS}, 33:\penalty0 6840--6851, 2020.

\bibitem[Jeong et~al.(2024)Jeong, Kim, Choi, Lee, and Uh]{jeong2024visual}
Jaeseok Jeong, Junho Kim, Yunjey Choi, Gayoung Lee, and Youngjung Uh.
\newblock Visual style prompting with swapping self-attention.
\newblock \emph{arXiv preprint}, 2024.

\bibitem[Kwon et~al.(2023)Kwon, Jeong, and Uh]{kwon2022diffusion}
Mingi Kwon, Jaeseok Jeong, and Youngjung Uh.
\newblock Diffusion models already have a semantic latent space.
\newblock In \emph{The Eleventh International Conference on Learning Representations}, 2023.

\bibitem[Lin et~al.(2024)Lin, Wang, and Yang]{lin2024sdxl}
Shanchuan Lin, Anran Wang, and Xiao Yang.
\newblock Sdxl-lightning: Progressive adversarial diffusion distillation.
\newblock \emph{arXiv preprint}, 2024.

\bibitem[Lu(2024)]{lu2024procreate}
Jack Lu.
\newblock Procreate, don’t reproduce! propulsive energy diffusion for creative generation.
\newblock In \emph{ECCV}, 2024.

\bibitem[Mateja and Heinzl(2021)]{mateja2021towards}
Deborah Mateja and Armin Heinzl.
\newblock Towards machine learning as an enabler of computational creativity.
\newblock \emph{IEEE Transactions on Artificial Intelligence}, 2\penalty0 (6):\penalty0 460--475, 2021.

\bibitem[Murray et~al.(2012)Murray, Marchesotti, and Perronnin]{murray2012ava}
Naila Murray, Luca Marchesotti, and Florent Perronnin.
\newblock Ava: A large-scale database for aesthetic visual analysis.
\newblock In \emph{CVPR}, pages 2408--2415, 2012.

\bibitem[Nobari et~al.(2021)Nobari, Rashad, and Ahmed]{nobari2021creativegan}
AH Nobari, MF Rashad, and F Ahmed.
\newblock Creativegan: Editing generative adversarial networks for creative design synthesis.
\newblock In \emph{IDETC-CIE}. American Society of Mechanical Engineers (ASME), 2021.

\bibitem[Podell et~al.(2024)Podell, English, Lacey, Blattmann, Dockhorn, M{\"u}ller, Penna, and Rombach]{podell2023sdxl}
Dustin Podell, Zion English, Kyle Lacey, Andreas Blattmann, Tim Dockhorn, Jonas M{\"u}ller, Joe Penna, and Robin Rombach.
\newblock Sdxl: Improving latent diffusion models for high-resolution image synthesis.
\newblock In \emph{ICLR}, 2024.

\bibitem[Razzhigaev et~al.(2023)Razzhigaev, Shakhmatov, Maltseva, Arkhipkin, Pavlov, Ryabov, Kuts, Panchenko, Kuznetsov, and Dimitrov]{razzhigaev2023kandinsky}
Anton Razzhigaev, Arseniy Shakhmatov, Anastasia Maltseva, Vladimir Arkhipkin, Igor Pavlov, Ilya Ryabov, Angelina Kuts, Alexander Panchenko, Andrey Kuznetsov, and Denis Dimitrov.
\newblock Kandinsky: An improved text-to-image synthesis with image prior and latent diffusion.
\newblock In \emph{EMNLP (Demos)}, 2023.

\bibitem[Richardson et~al.(2024)Richardson, Goldberg, Alaluf, and Cohen-Or]{richardson2023conceptlab}
Elad Richardson, Kfir Goldberg, Yuval Alaluf, and Daniel Cohen-Or.
\newblock Conceptlab: Creative concept generation using vlm-guided diffusion prior constraints.
\newblock \emph{ACM Transactions on Graphics}, 43\penalty0 (3):\penalty0 1--14, 2024.

\bibitem[Rombach et~al.(2022)Rombach, Blattmann, Lorenz, Esser, and Ommer]{rombach2022high}
Robin Rombach, Andreas Blattmann, Dominik Lorenz, Patrick Esser, and Bj{\"o}rn Ommer.
\newblock High-resolution image synthesis with latent diffusion models.
\newblock In \emph{CVPR}, pages 10684--10695, 2022.

\bibitem[Sauer et~al.(2025)Sauer, Lorenz, Blattmann, and Rombach]{sauer2025adversarial}
Axel Sauer, Dominik Lorenz, Andreas Blattmann, and Robin Rombach.
\newblock Adversarial diffusion distillation.
\newblock In \emph{ECCV}, pages 87--103. Springer, 2025.

\bibitem[Sbai et~al.(2018)Sbai, Elhoseiny, Bordes, LeCun, and Couprie]{sbai2018design}
Othman Sbai, Mohamed Elhoseiny, Antoine Bordes, Yann LeCun, and Camille Couprie.
\newblock Design: Design inspiration from generative networks.
\newblock In \emph{ECCV workshops}, pages 37--44, 2018.

\bibitem[Si et~al.(2024)Si, Huang, Jiang, and Liu]{si2024freeu}
Chenyang Si, Ziqi Huang, Yuming Jiang, and Ziwei Liu.
\newblock Freeu: Free lunch in diffusion u-net.
\newblock In \emph{CVPR}, pages 4733--4743, 2024.

\bibitem[Song et~al.(2021)Song, Meng, and Ermon]{songdenoising}
Jiaming Song, Chenlin Meng, and Stefano Ermon.
\newblock Denoising diffusion implicit models.
\newblock In \emph{ICLR}, 2021.

\bibitem[Vinker et~al.(2023)Vinker, Voynov, Cohen-Or, and Shamir]{vinker2023concept}
Yael Vinker, Andrey Voynov, Daniel Cohen-Or, and Ariel Shamir.
\newblock Concept decomposition for visual exploration and inspiration.
\newblock \emph{ACM TOG}, 42\penalty0 (6):\penalty0 1--13, 2023.

\bibitem[Voynov et~al.(2023)Voynov, Chu, Cohen-Or, and Aberman]{voynov2023p+}
Andrey Voynov, Qinghao Chu, Daniel Cohen-Or, and Kfir Aberman.
\newblock p+: Extended textual conditioning in text-to-image generation.
\newblock \emph{arXiv preprint}, 2023.

\bibitem[Wang et~al.(2024)Wang, Chen, and Wang]{wang2024diffusionbasedvisualartcreation}
Bingyuan Wang, Qifeng Chen, and Zeyu Wang.
\newblock Diffusion-based visual art creation: A survey and new perspectives.
\newblock \emph{arXiv preprint arXiv:2408.12128}, 2024.

\bibitem[Xu et~al.(2024)Xu, Corso, Jaakkola, Vahdat, and Kreis]{xu2024disco}
Yilun Xu, Gabriele Corso, Tommi Jaakkola, Arash Vahdat, and Karsten Kreis.
\newblock Disco-diff: enhancing continuous diffusion models with discrete latents.
\newblock In \emph{ICML}, pages 54933--54961, 2024.

\bibitem[Yang et~al.(2024)Yang, Zhang, Song, Hong, Xu, Zhao, Zhang, Cui, and Yang]{yang2024diffusionmodelscomprehensivesurvey}
Ling Yang, Zhilong Zhang, Yang Song, Shenda Hong, Runsheng Xu, Yue Zhao, Wentao Zhang, Bin Cui, and Ming-Hsuan Yang.
\newblock Diffusion models: A comprehensive survey of methods and applications, 2024.

\bibitem[Yu et~al.(2023)Yu, Wang, Zhao, Ghanem, and Zhang]{yu2023freedomtrainingfreeenergyguidedconditional}
Jiwen Yu, Yinhuai Wang, Chen Zhao, Bernard Ghanem, and Jian Zhang.
\newblock Freedom: Training-free energy-guided conditional diffusion model, 2023.

\bibitem[Zhang et~al.(2023)Zhang, Rao, and Agrawala]{zhang2023adding}
Lvmin Zhang, Anyi Rao, and Maneesh Agrawala.
\newblock Adding conditional control to text-to-image diffusion models.
\newblock In \emph{ICCV}, pages 3836--3847, 2023.

\end{thebibliography}
}

\appendix

\clearpage
\counterwithin{figure}{section}
\renewcommand{\thefigure}{\Alph{section}.\arabic{figure}}
\renewcommand{\thetable}{\Alph{table}}
\renewcommand{\thesection}{\Alph{section}}
\setcounter{page}{1}
\onecolumn
\maketitlesupplementary
%
%

\section{Implementation Details}
\label{sec:app_settings}


\textbf{Baseline Settings.} We compared our \textbf{C3} method with the original Stable Diffusion-based models and ConceptLab~\cite{richardson2023conceptlab}. Specifically, we evaluated four Stable Diffusion-based models: SDXL and its distilled variants—Turbo, Lightning 1-step, and Lightning 4-step. The results for each model are compared with the corresponding version enhanced by \textbf{C3}. Unless specified otherwise, we used the default settings of the original models, including the classifier-free guidance scale and negative prompts, to ensure a consistent baseline for evaluating the effectiveness of the \textbf{C3} method.
For the ConceptLab, we adhered to the default settings outlined in the original paper. These settings include a batch size of 1, 2500 training steps, and ``[object]" as the positive class. To generate 100 samples, we trained the embeddings using 10 different initial seeds, and for each trained embedding, we generated 10 new images during inference.

\noindent\textbf{C3 Settings.} In Table~\ref{tab:appx_settings}, we provide the detailed parameter settings used in Section~\ref{sec:experiment}. These settings are designed to be broadly applicable, ensuring that configuring parameters within the provided range will likely produce satisfactory results for most prompts. 
The cut-off parameters indicate the extent to which specific frequencies are amplified. For the Turbo model, we set the cut-off threshold to 5 for every block. For the other models, the cut-off thresholds were set as [10, 5, 5, 5], corresponding to their respective blocks. The rationale for these settings is that the cut-off threshold should align with the resolution of the internal features, ensuring optimal handling of feature granularity at different blocks.
For the amplification factor selection, we use the mean usability score $\text{Use}(\mathcal{I}) = \frac{1}{N}\sum_{i=1}^N Aesthetic(I_i)+\frac{1}{N}\sum_{i=1}^N CLIP(I_i,c)$ for $\mathcal{I}=\{I_i\}_1^N$ to provide statistically consistent amplification factors. In the experiments, we use the number of samples $N=100$. To balance the scale of the aesthetic score and CLIP score, we min-max scale each score over the configurations $\{(l,   \lambda_l^i )\}_{(l, i)}$. 
The usability bumper is a parameter designed to balance the usability and novelty of the images generated with \textbf{C3}. For the SDXL model, we set the usability bumper to 0.7, while for the other models, it was set to 0.8.
For the sum constraint applied to scaling factors, which controls the degree of amplification across multiple blocks, we used the sum values of 0.6, 0.8, and 1. The specific value was chosen based on the given prompt and the model in use. Additionally, we provide detailed block-wise scaling factors for a more comprehensive understanding of the amplification strategy.
In the next section, we conduct in-depth analyses of the effects of various hyperparameters on the results.

\begin{table*}[!ht]
\centering
\fontsize{9.5}{13}\selectfont
\begin{tabular}{cc|c|c|cccc|l}
\toprule \toprule
\multicolumn{2}{c|}{}                                                               & Cut-off & Usability Bumper  & \multicolumn{4}{c|}{Amplification Factors} & Block-wise Scaling Factors  \\ \hline
\multicolumn{1}{c|}{\multirow{5}{*}{SDXL}}                                           & chair  & {[}10,5,5,5{]} & 0.7 &1.15&1.6&5&6 
 & sum{[}0.3,0.3,0.1,0.1{]}=0.8 \\  \cline{2-9} 
\multicolumn{1}{c|}{}                                                               & teddy bear & {[}10,5,5,5{]} & 0.7 &1.2&1.8&4&2 
 & sum{[}0.4,0.4,0.1,0.1{]}=1.0  \\ \cline{2-9} 
\multicolumn{1}{c|}{}                                                               & car & {[}10,5,5,5{]} & 0.7 &1.25&1.6&5&4 
 & sum{[}0.3,0.3,0.1,0.1{]}=0.8 \\  \cline{2-9} 
\multicolumn{1}{c|}{}                                                               & building  & {[}10,5,5,5{]} & 0.7 & 1.25&1.8&5&4 
 & sum{[}0.2,0.15,0.15,0.1{]}=0.6  \\ \cline{2-9} 
\multicolumn{1}{c|}{}                                                               & garment    & {[}10,5,5,5{]} & 0.7 & 1.2&1.8&5&2 
  & sum{[}0.4,0.4,0.1,0.1{]}=1.0 \\ \bottomrule
\multicolumn{1}{c|}{\multirow{6}{*}{\begin{tabular}[c]{@{}c@{}}Lightning\ (1-step)\end{tabular}}} & chair & {[}10,5,5,5{]} & 0.8  & 1.5&2.25&5&6

  & sum{[}0.2,0.2,0.1,0.1{]}=0.6 \\ \cline{2-9} 
\multicolumn{1}{c|}{}                                                               & teddy bear & {[}10,5,5,5{]} & 0.8  & 1.5&2.75&6&7

 & sum{[}0.2,0.2,0.1,0.1{]}=0.6 \\ \cline{2-9} 
\multicolumn{1}{c|}{}                                                               & car & {[}10,5,5,5{]} & 0.8  & 1.5&2.5&6&6

 & sum{[}0.2,0.2,0.1,0.1{]}=0.6 \\  \cline{2-9} 
\multicolumn{1}{c|}{}                                                               & building   & {[}10,5,5,5{]} & 0.8 & 1.6&2.5&7&8

 & sum{[}0.2,0.2,0.1,0.1{]}=0.6  \\ \cline{2-9} 
\multicolumn{1}{c|}{}                                                               & garment    & {[}10,5,5,5{]} & 0.8 & 1.3&1.9&6&7

 & sum{[}0.2,0.2,0.1,0.1{]}=0.6 \\  \cline{2-9} 
\multicolumn{1}{c|}{}                                                               & fish       & {[}10,5,5,5{]} & 0.8  & 1.4&2.75&5&5

 & sum{[}0.2,0.2,0.1,0.1{]}=0.6 \\ \bottomrule
\multicolumn{1}{c|}{\multirow{5}{*}{\begin{tabular}[c]{@{}c@{}}Lightning\ (4-step)\end{tabular}}} & chair      & {[}10,5,5,5{]} & 0.8 & 1.4&2&7&8

 & sum{[}0.2,0.15,0.15,0.1{]}=0.6 \\ \cline{2-9} 
\multicolumn{1}{c|}{}                                                               & teddy bear & {[}10,5,5,5{]} & 0.8  & 1.4&2.25&6&9

 & sum{[}0.2,0.15,0.15,0.1{]}=0.6 \\ \cline{2-9} 
\multicolumn{1}{c|}{}                                                               & car        & {[}10,5,5,5{]} & 0.8  & 1.4&1.9&6&6

 & sum{[}0.2,0.15,0.15,0.1{]}=0.6\\ \cline{2-9} 
\multicolumn{1}{c|}{}                                                               & building   & {[}10,5,5,5{]} & 0.8  & 1.3&1.9&8&7

 & sum{[}0.2,0.15,0.15,0.1{]}=0.6\\ \cline{2-9} 
\multicolumn{1}{c|}{}                                                               & garment    & {[}10,5,5,5{]} & 0.8 & 1.25&1.8&5&4

  & sum{[}0.2,0.15,0.15,0.1{]}=0.6\\ \bottomrule
\multicolumn{1}{c|}{\multirow{6}{*}{Turbo}}                                    & chair      & {[}5,5,5,5{]}  & 0.8  & 1.25&1.5&9&10

  & sum{[}0.3,0.3,0.2,0.2{]}=1.0 \\ \cline{2-9} 
\multicolumn{1}{c|}{}                                                               & teddy bear & {[}5,5,5,5{]}  & 0.8 & 2&1.5&7&10

  & sum{[}0.4,0.4,0.1,0.1{]}=1.0 \\  \cline{2-9} 
\multicolumn{1}{c|}{}                                                               & car        & {[}5,5,5,5{]}  & 0.8 & 1.75&2.5&8&10

  & sum{[}0.4,0.4,0.1,0.1{]}=1.0 \\  \cline{2-9} 
\multicolumn{1}{c|}{}                                                               & building   & {[}5,5,5,5{]}  & 0.8 & 2.75&3.75&10&10

 & sum{[}0.2,0.15,0.15,0.1{]}=0.6 \\ \cline{2-9} 
\multicolumn{1}{c|}{}                                                               & garment    & {[}5,5,5,5{]}  & 0.8  & 1.5&2.25&7&10

 & sum{[}0.4,0.4,0.1,0.1{]}=1.0 \\  \cline{2-9} 
\multicolumn{1}{c|}{}                                                               & sunglasses & {[}5,5,5,5{]}  & 0.8  & 3.75&5&7&10 & sum{[}0.1,0.1,0.2,0.2{]}=0.6 \\ \bottomrule \bottomrule
\end{tabular}
\caption{Detailed Settings for the used parameters in experiments. The numbers within the list represent the corresponding values applied to each block.}
\label{tab:appx_settings}
\end{table*}

\section{Ablation Study on the Hyperparameters}
\label{sec:app_hyperparameter}
\subsection{Analysis on Cutoff Threshold}
\begin{figure*}
     \centering
     \begin{subfigure}[htbp]{0.95\textwidth}
         \centering
         \includegraphics[width=\textwidth]{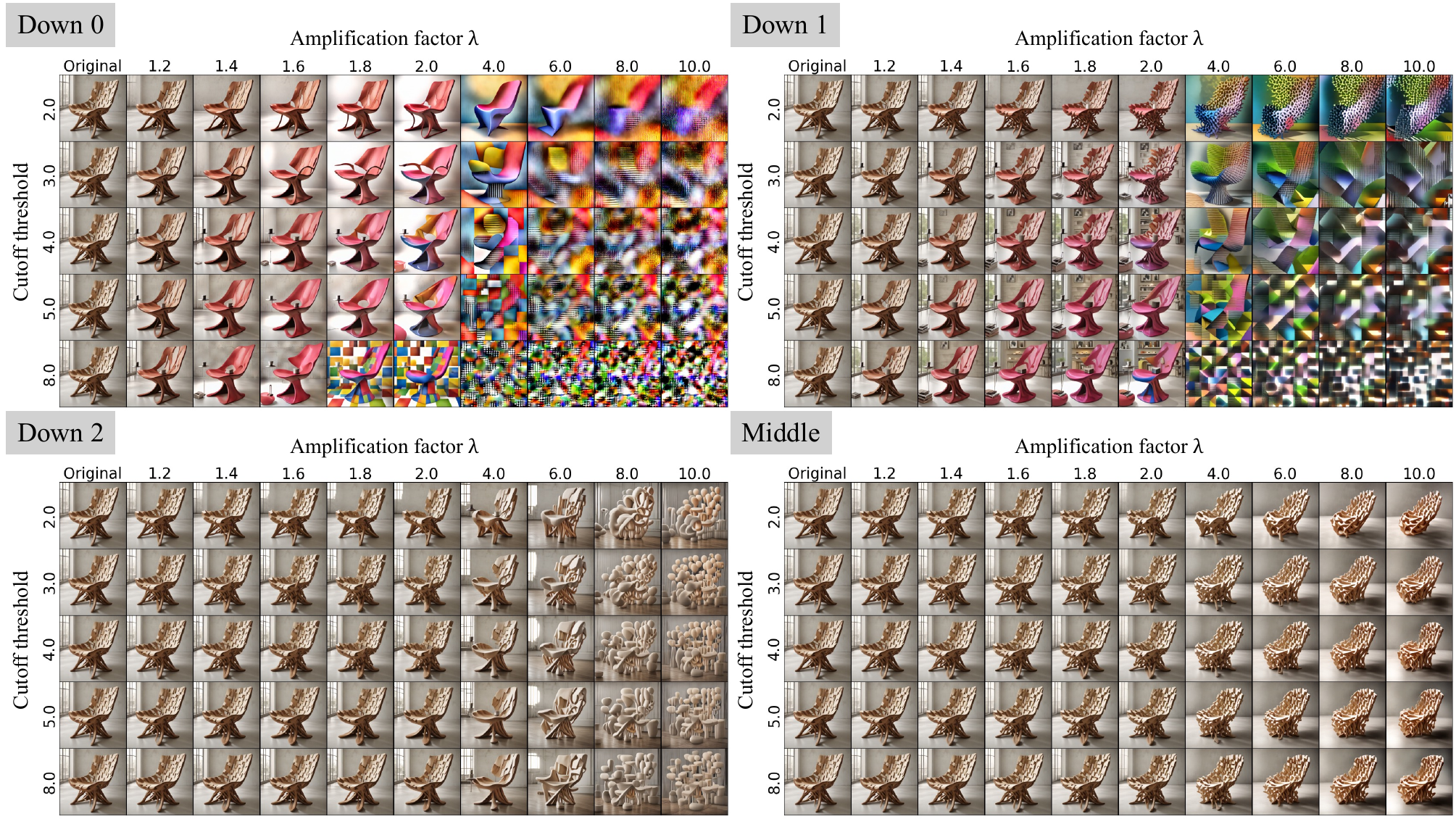}
         \caption{SDXL-Turbo, prompt=``a creative chair"}
     \end{subfigure}
     \hfill
     \vskip 1cm
     \begin{subfigure}[htbp]{0.95\textwidth}
         \centering
         \includegraphics[width=\textwidth]{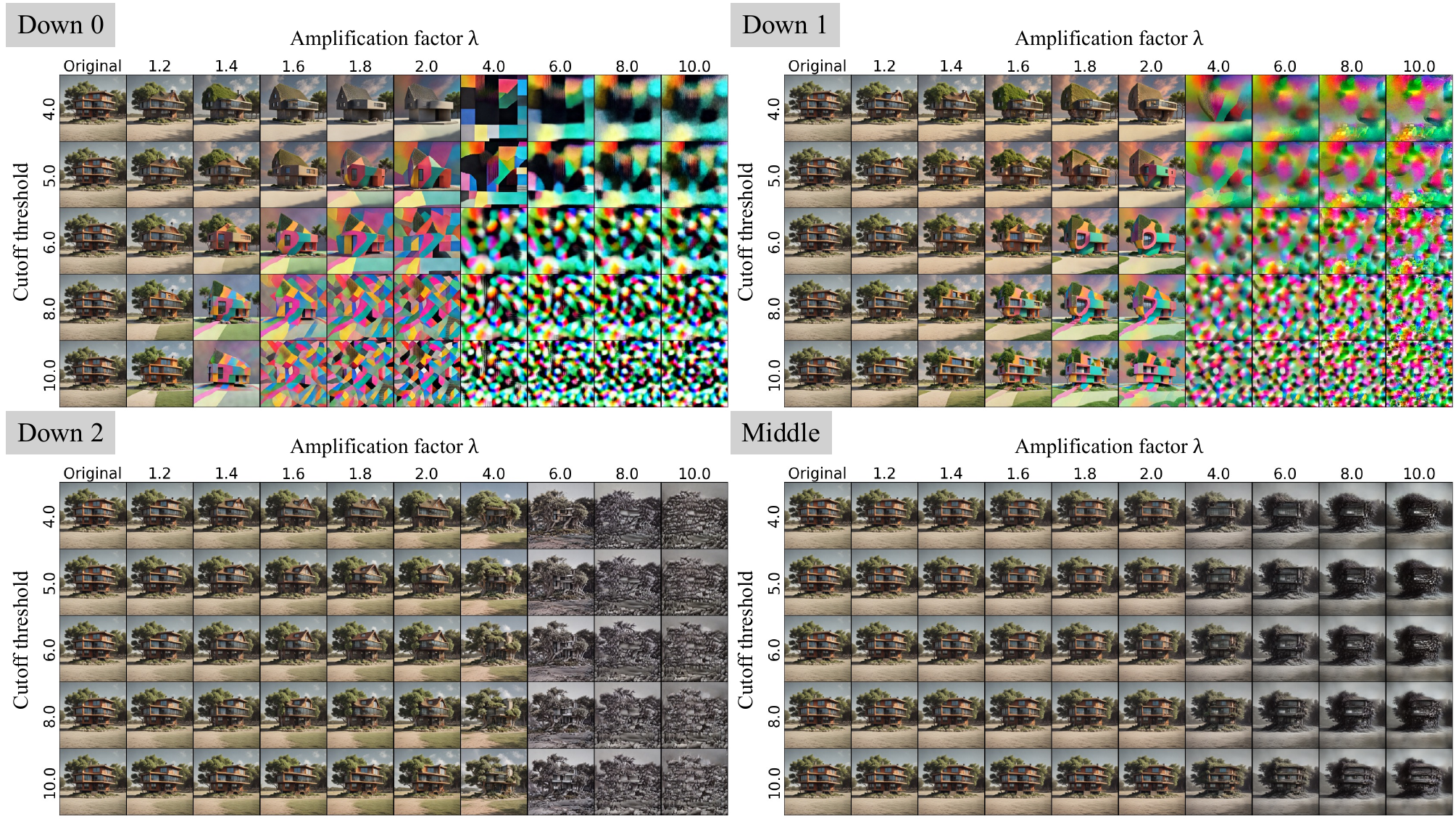}
         \caption{SDXL-Lightning (1-step), prompt=``a creative house"}
     \end{subfigure}
        \caption{Amplification results for various cutoff thresholds.}
        \label{fig:app_cutoff}
\end{figure*}

In this subsection, we analyze the effect of various cutoff thresholds. The cutoff threshold $c$ defines the extent to which frequency we would amplify. The low-frequency mask $M_L\in[0,1]^{n\times n}$ is then defined with the cutoff threshold $c$ as follows.
\begin{equation}
    M_L^{i,j} = \begin{cases}
        1 \quad \text{ if  } r(i,j) < c \\
        0 \quad otherwise
    \end{cases}
\end{equation}
Here, $M_L^{i,j}$ denotes the element of $M_L$ located at the $i^{th}$ row and $j^{th}$ column, and $r(i,j)=\sqrt{(i-\frac{n}{2})^2+(j-\frac{n}{2})^2}$. The resolution $n \times n$ varies across blocks and models. Therefore, the resolution should be considered when setting the cutoff threshold.
In Figure \ref{fig:app_cutoff}, we present the detailed amplification results for various cutoff thresholds for each block. In the first and second down blocks, a larger cutoff threshold facilitates more colorful variation. However, too large cutoff threshold introduces a tile pattern in the image that degrades quality. Conversely, a smaller cutoff threshold successfully prevents this tile pattern but, if set too low, can result in excessive information loss and over-smoothing of the object. By adjusting the cutoff threshold, one can find outcomes with a unique shape and color pattern. Compared to the shallow blocks, the amplification results on the third down block and the middle block indicate that these deeper blocks are less sensitive to the cutoff thresholds. 
Furthermore, we observe that a smaller cutoff threshold generally permits greater amplification, while a larger cutoff threshold tends to generate noise images with a smaller amplification factor. 

\subsection{Analysis on Usability Bumper}
\begin{figure*}[th]
  \centering
  
         \includegraphics[width=\textwidth]{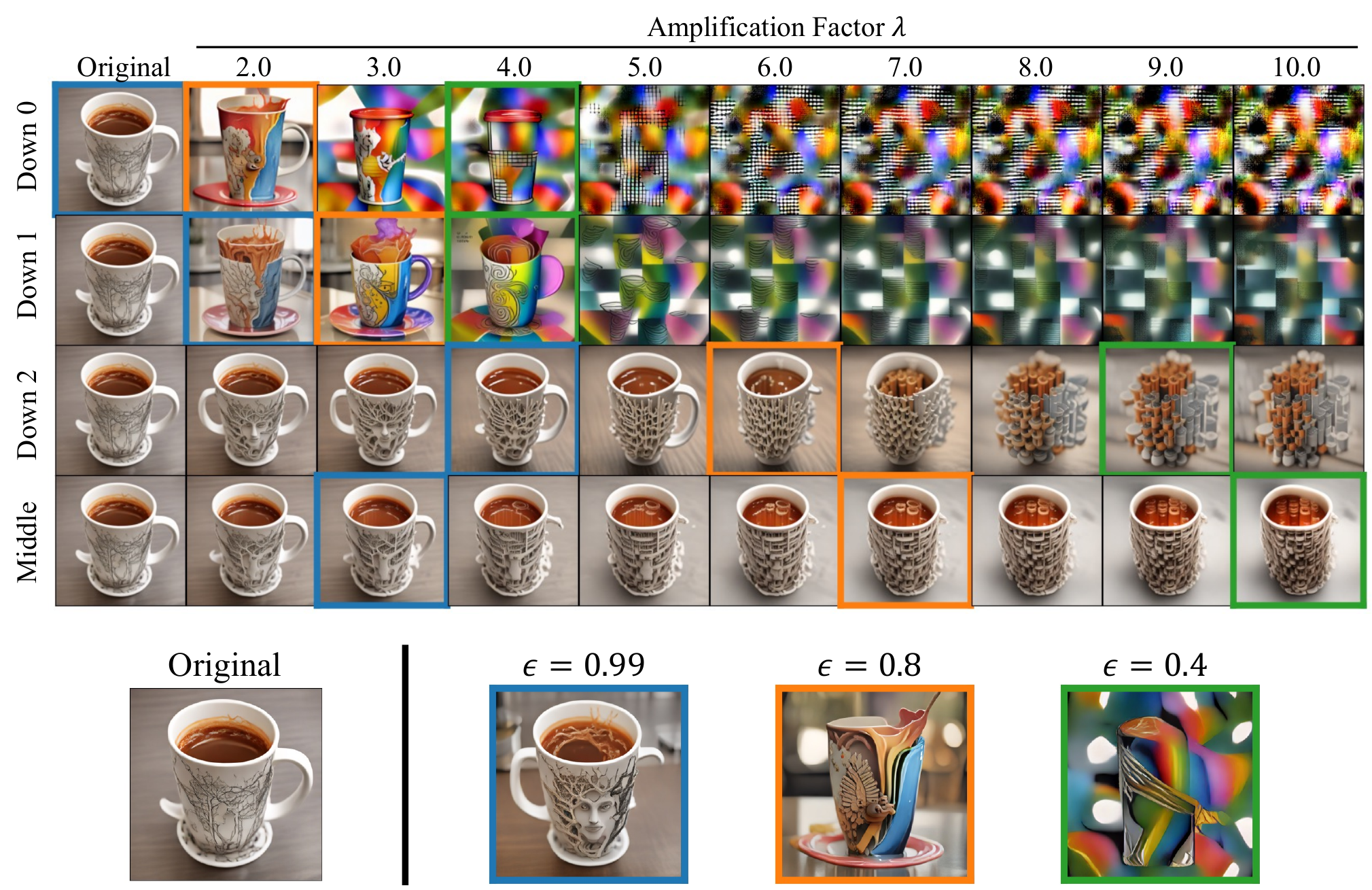}
         \caption{ The amplified results with the various usability buffer $\epsilon$. 
         }
         \label{fig:app_usability}
  
\end{figure*}

In this subsection, we analyze the effect of the usability bumper $\epsilon$ defined in Section \ref{sec:alg_params}. The usability bumper is used as a control parameter between usability and novelty. When $\epsilon$ is close to 1.0, the usability score is preserved similar to that of the original image, albeit with a loss of novelty. Conversely, as $\epsilon$ decreases, it permits greater variation from the original image and enhances novelty, albeit at the expense of the usability score. Figure \ref{fig:app_usability} shows examples with the use of the various usability bumper. Turbo with the prompt ``a creative cup" is used for the generation. The images with the colored bounding boxes indicate the selected amplification factors with Equation (\ref{eq:usability}) in Section \ref{sec:alg_params}. For each block, the amplification factors found with $\epsilon=0.99$, marked with the blue bounding boxes, produce images that maintain high fidelity to the original image, with only slight changes in detail. Conversely, the amplification factors identified with $\epsilon = 0.4$, indicated by the green bounding boxes, generate images with significant variation from the original. Specifically, the shallower blocks exhibit artistic cup images with high color variation, while the deeper blocks show changes primarily in shape, albeit with compromised image fidelity. The amplification factors identified with $\epsilon = 0.8$ as used in the main experiment, marked with the orange bounding boxes, result in outcomes that fall between these two extremes. The following three images, indicated by colored bounding boxes, display the results of amplification across all four blocks. For the scaling factors, 0.3, 0.3, 0.2, 0.2 are used for each block, respectively. The results indicate that $\epsilon = 0.8$ produces a cup image that is both creative and feasible.  
\vskip -0.5cm

\subsection{Analysis on Scaling Factors}

We introduce an automatic strategy to determine the optimal amplification factor $\lambda^*_l$ in Section~\ref{sec:alg_params}. This approach strikes a balance between usability and novelty, generating semantically meaningful yet creative features that lead to creative images.

Applying our method across multiple blocks simultaneously enables the generation of more flexible and creative images. However, when the changes in multiple blocks are simply accumulated, the resulting features may exceed the allowable range of the pre-trained Stable Diffusion-based model, leading to broken or degraded images. To address this, we apply additional scaling factors $s_l$ during multi-block applications of \textbf{C3} to preserve image quality. Then, we can formulate the \textbf{C3} method applied across multi-blocks as follows: 
 
\begin{equation}
    f^*(x_l) = s_{l} \cdot\lambda_l^* \cdot f_L(x_l) + f_H(x_l)
\end{equation}

Empirically, we observe that bounding the sum of scaling factors across the blocks, denoted as $S=\sum_l s_l$, aids in preventing extensive parameter search for $s_l$. Within this sum constraint, the block-specific scaling factors can be adjusted in a user-controllable manner, allowing for flexible image generation. We observed that selecting an appropriate sum constraint prevents degradation in image quality, even as the scaling factors for individual blocks vary. (See Figure~\ref{fig:app_scale_factor_quan})

\begin{figure*}[!th]
  \centering
  
         \includegraphics[width=0.9\textwidth]{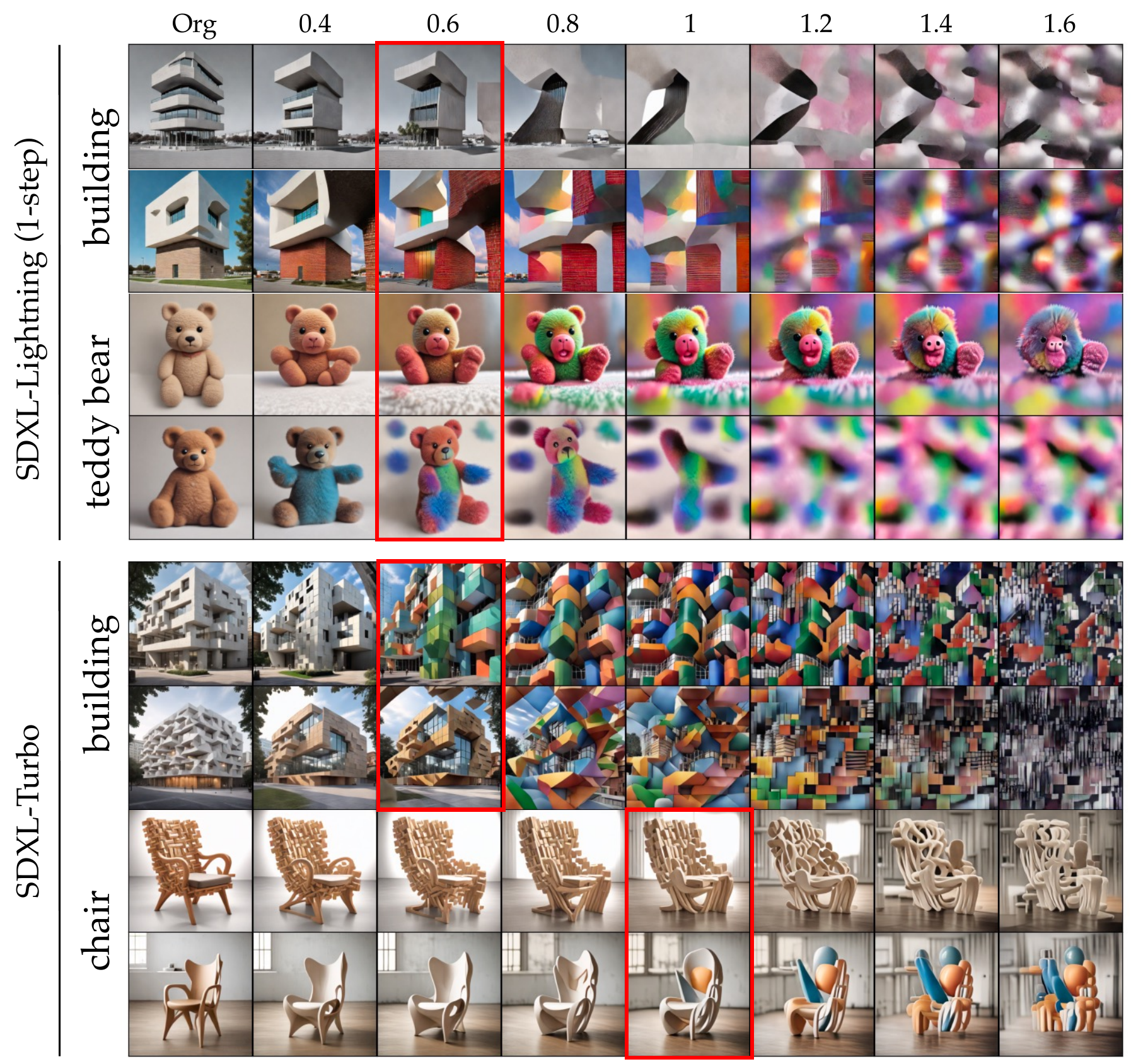}
         \caption{Amplification results for various scaling factors with the sum constraint. The red box presents the sum value we used for each object. }
         \label{fig:app_scale_factor}
  
\end{figure*}

In Figure~\ref{fig:app_scale_factor}, we display the variations in images for each model and object as $S$ changes. Results highlighted in red boxes represent the constraint value we used. (Specific block-wise scaling factor settings for the figures are summarized in Table~\ref{tab:appx_settings}). For the SDXL-Lightning 1-step model, we observe that using a summation constraint of $S=0.6$ resulted in the most creative images while maintaining the usability of the object in most cases. For the SDXL-Turbo model, a summation constraint of approximately $S=1$ produces highly creative results while effectively preserving the structural integrity of objects like chairs. However, for more complex objects, such as buildings, the cumulative amplification tends to introduce additional noise, requiring a more conservative summation constraint to balance creativity and object clarity.


Furthermore, we quantitatively analyze the necessity of the scaling factor and its correlation with usability, which is measured using the BLIP score. As introduced earlier in Section~\ref{sec:quan}, BLIP score represents the proportion of generated samples that receive a ``yes" response from the BLIP VQA model when asked, ``Is this image [object]?". For each sum of scaling factors, 100 images are generated with different scaling factors. These 100 cases were obtained by randomly sampling scaling factors for each block,$s_l$, within the given sum value.

The results reveal that as the sum of the scaling factors increases, the BLIP score decreases, indicating that larger scaling factors compromise usability. We set the scaling factor constraints to a value that ensures the model does not compromise its usability significantly, as highlighted with bold markers. Importantly, these findings are based on randomly selected scaling factors, demonstrating that the quality of the generated images remains robust within the specified sum constraint, regardless of how the scaling factors are distributed across blocks.

\begin{figure*}[!th]
  \centering
  
         \includegraphics[width=0.9\textwidth]{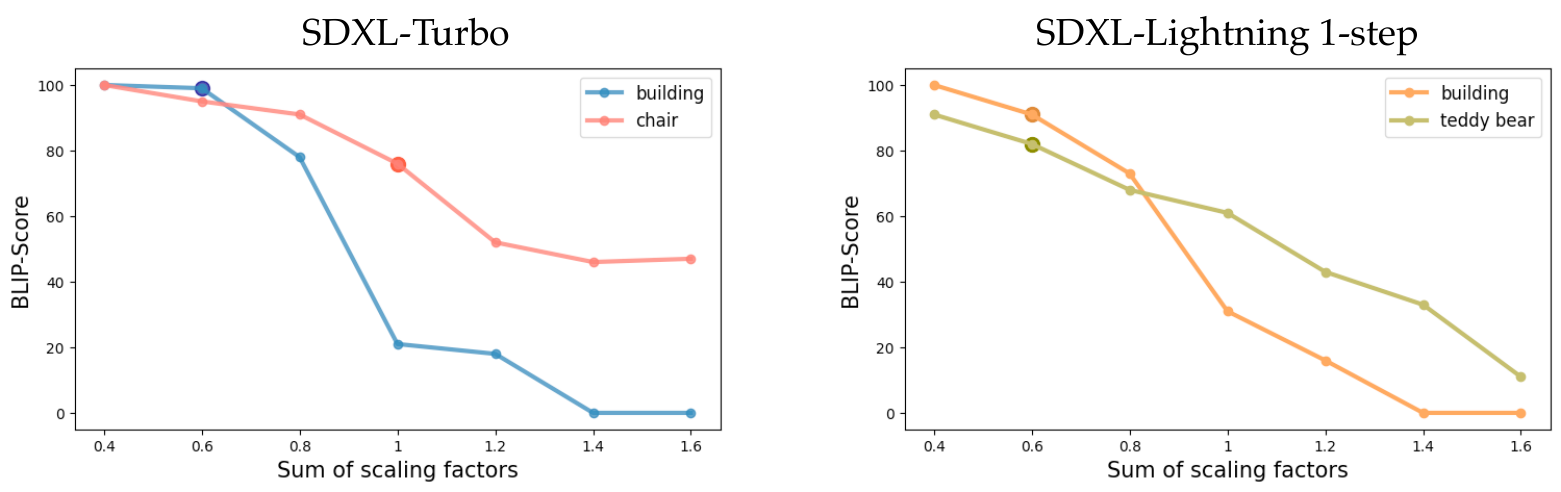}
         \caption{BLIP scores of various scaling factors.}
         \label{fig:app_scale_factor_quan}
  
\end{figure*}

\subsection{Step-wise Analysis}
\begin{figure}[!t]
  \centering
  
         \includegraphics[width=0.8\columnwidth]{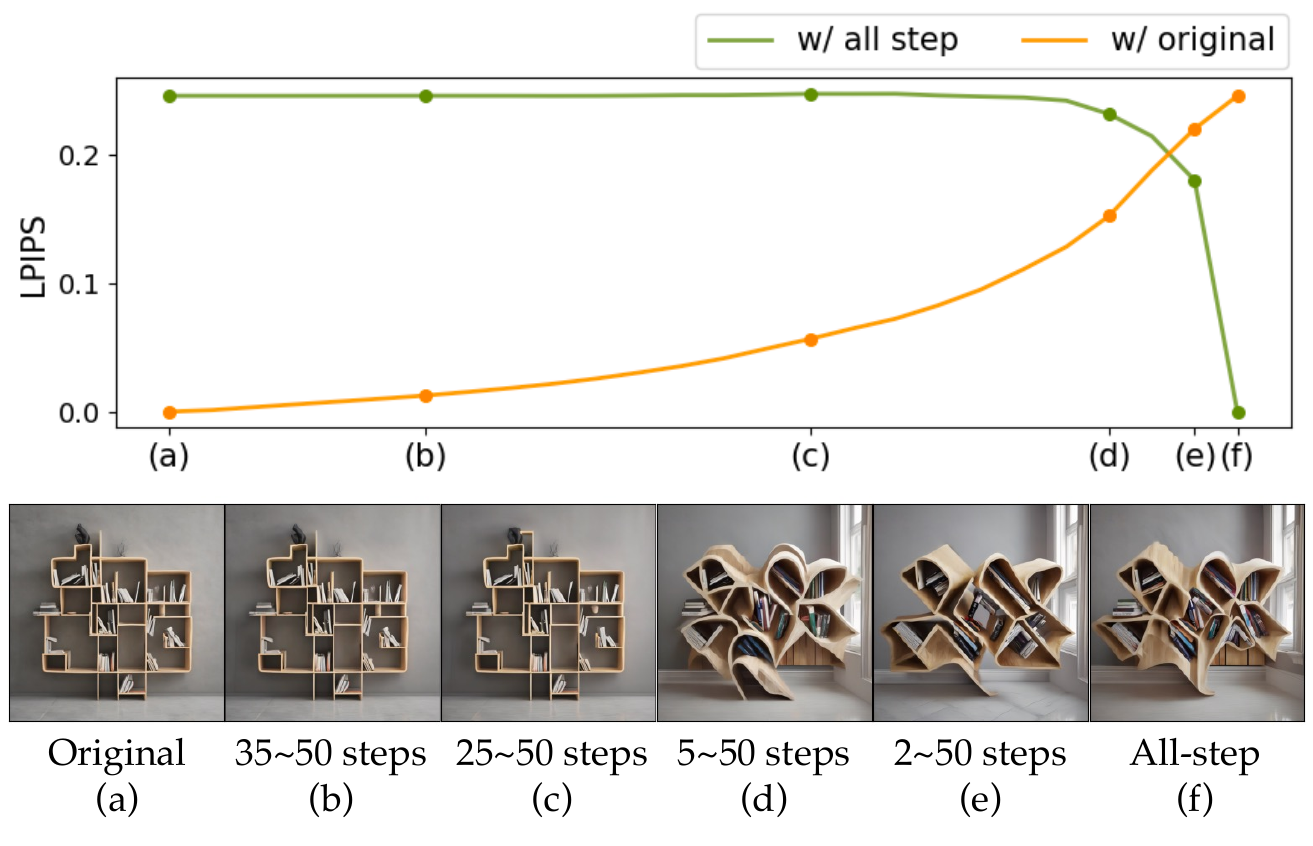}
         \caption{Step-wise amplification results on SDXL (step=50). Significant changes are observed in the earlier steps.  
         }
         \label{fig:step}
  
\end{figure}
Diffusion models operate through a multi-step denoising process. In this section, we examine the effects of applying C3 at various stages within this denoising process. We observe the changes in generated images by applying our method at six distinct points across a total of 50 steps. To quantify the degree of image change, we use LPIPS, a perceptual similarity metric, to compare the results at each stage with those generated by the proposed method. The LPIPS scores, averaged across 100 different images and random seeds, are shown in Figure~\ref{fig:step}-(top). Both the LPIPS scores and exemplar images show that when C3 is applied after the fifth step (Figure~\ref{fig:step}-(c)), the resulting images increasingly resemble the original, diverging from those generated with C3 applied continuously at each step. This analysis demonstrates that the impact of our method is most pronounced when applied in the early stages of the diffusion process, aligning with prior analyses on diffusion models that suggest structural content is primarily established in the earlier timesteps~\cite{yu2023freedomtrainingfreeenergyguidedconditional}.

\clearpage
\section{Detailed Experimental Results}
\label{sec:app_detailed_exp}

\subsection{Qualitative Results}
\subsubsection{Uncurated Samples for SDXL-Lightning (1-step)}
\begin{figure*}[htbp]
  \centering
  
         \includegraphics[width=\textwidth]{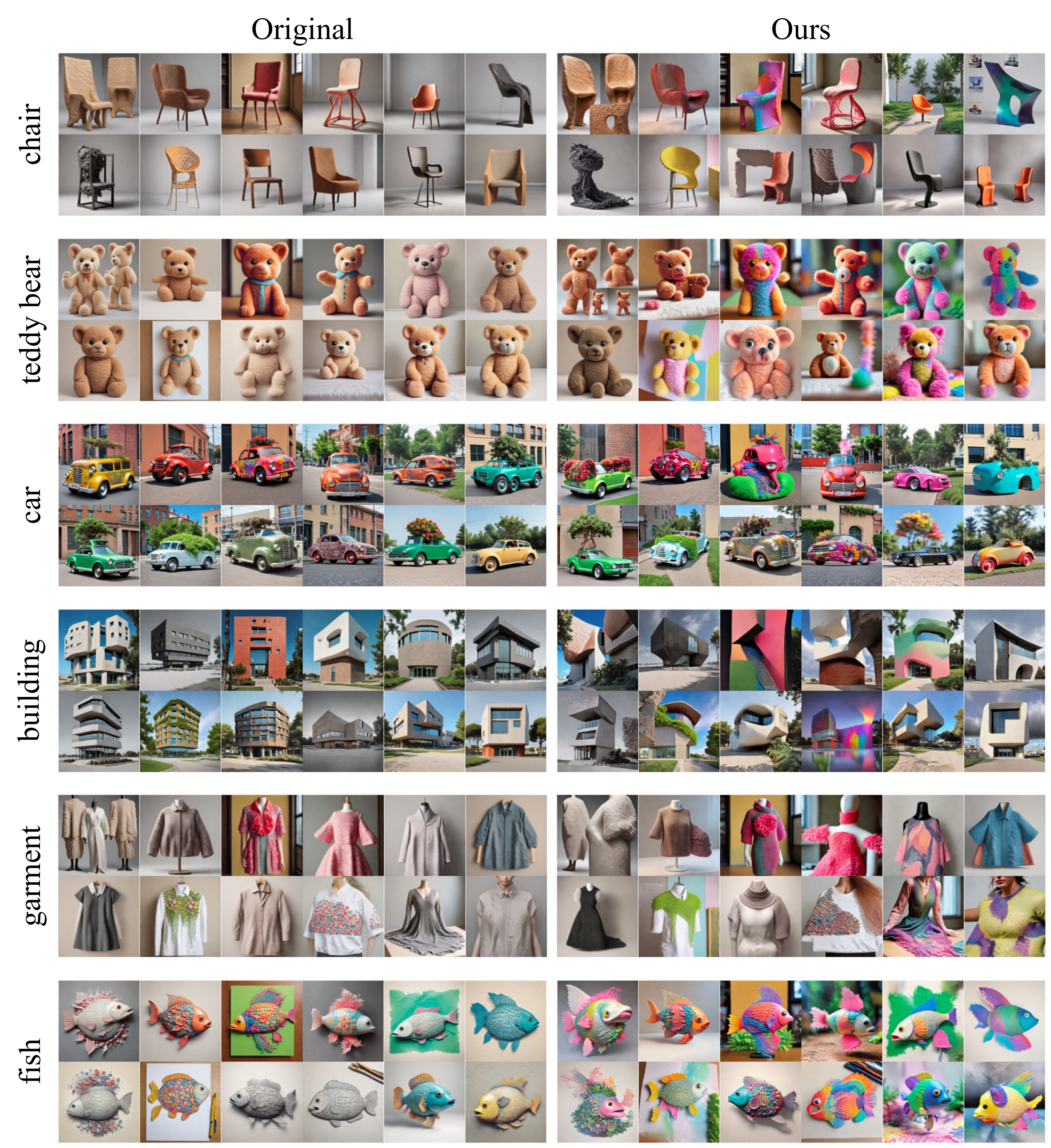}
         \caption{Uncurated samples generated from SDXL-Lightning (1-step). The samples are generated by manually setting the random seed to values ranging from 0 to 11.
         }
         \label{fig:app_qual_light1step}
  
\end{figure*}
\clearpage
\subsubsection{Uncurated Samples for SDXL-Turbo}
\begin{figure*}[htbp]
  \centering
  
         \includegraphics[width=\textwidth]{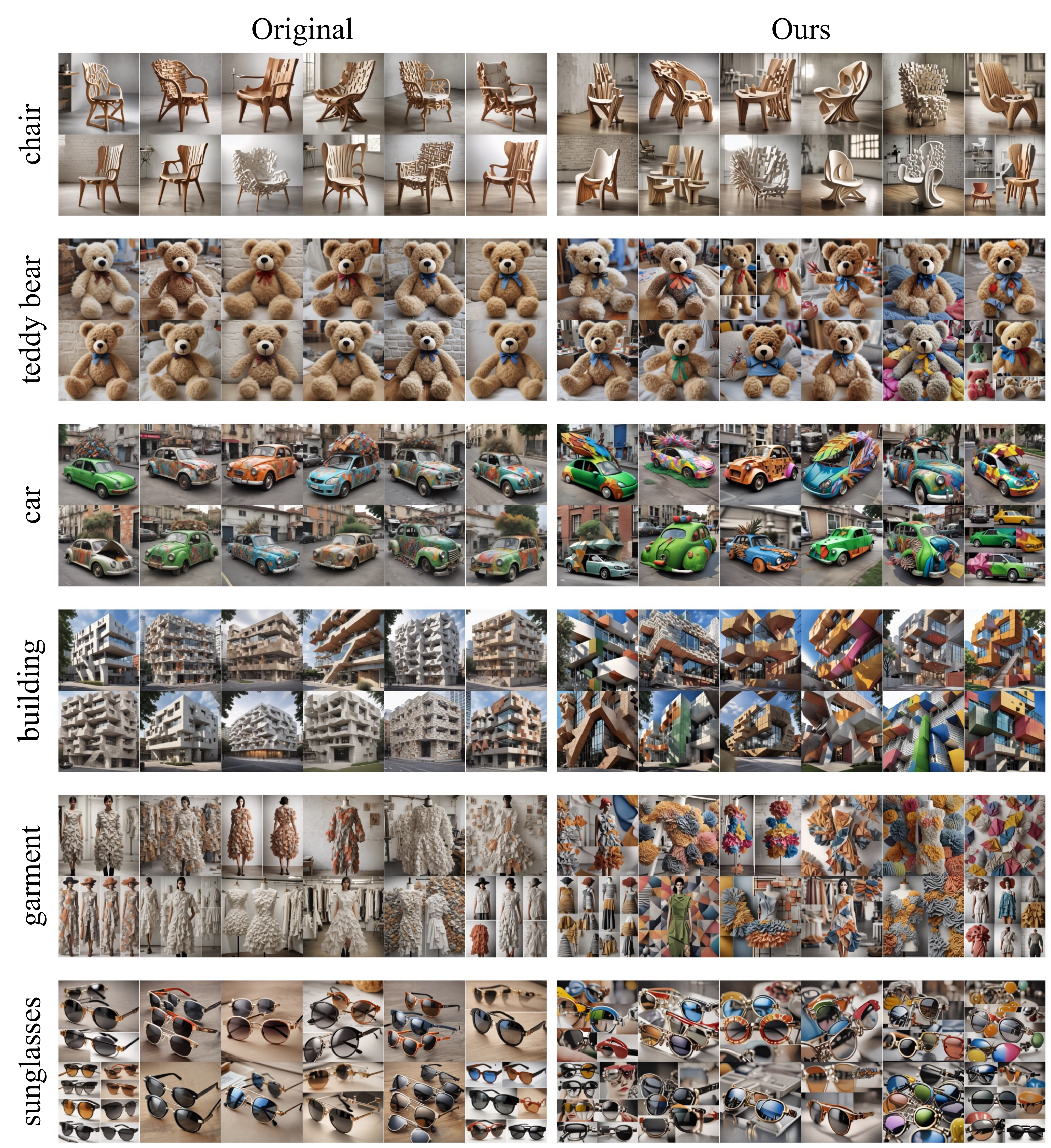}
         \caption{Uncurated samples generated from SDXL-Turbo. The samples are generated by manually setting the random seed to values ranging from 0 to 11.  
         }
         \label{fig:app_qual_turbo}
  
\end{figure*}
\clearpage
\subsubsection{Uncurated Samples for SDXL-Lightning (4-step)}
\begin{figure*}[htbp]
  \centering
  
         \includegraphics[width=0.9\textwidth]{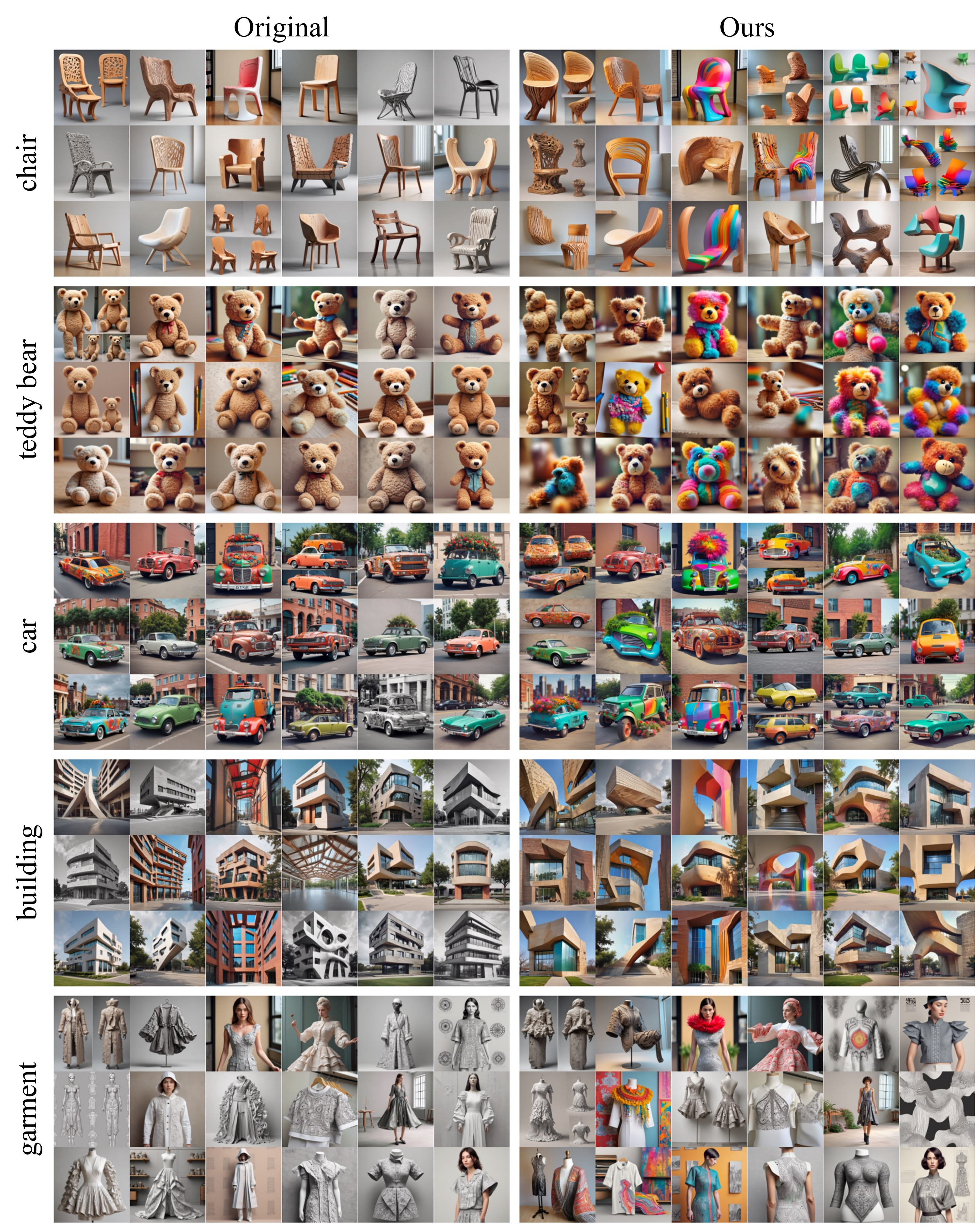}
         \caption{Uncurated samples generated from SDXL-Lightning (4-step). The samples are generated by manually setting the random seed to values ranging from 0 to 17.
         }
         \label{fig:app_qual_light4step}
  
\end{figure*}
\clearpage
\subsubsection{Uncurated Samples for SDXL}
\begin{figure*}[htbp]
  \centering
  
         \includegraphics[width=0.9\textwidth]{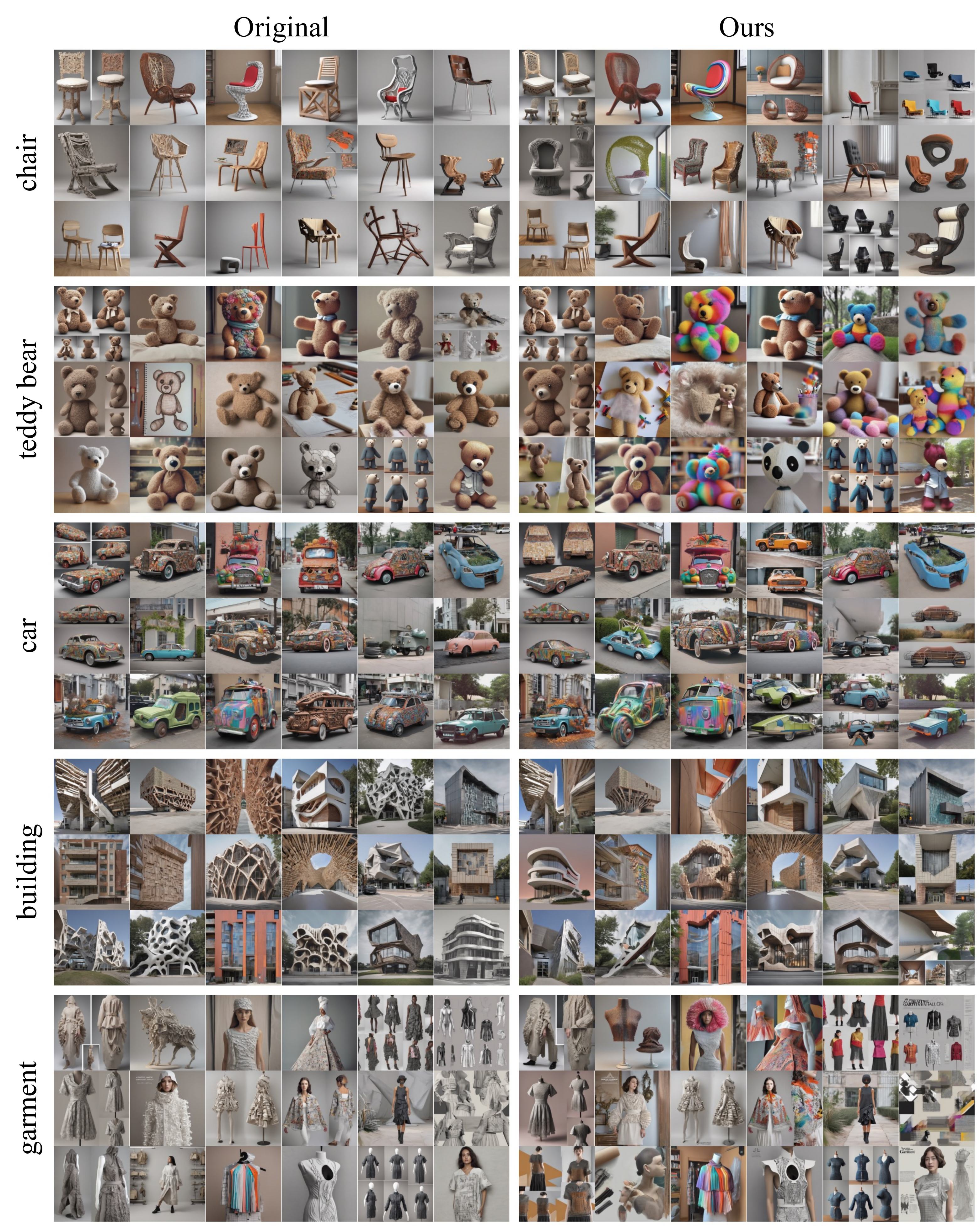}
         \caption{Uncurated samples generated from SDXL. The samples are generated by manually setting the random seed to values ranging from 0 to 17. 
         }
         \label{fig:app_qual_sdxl}
  
\end{figure*}
\clearpage
\subsubsection{Uncurated Samples for ConceptLab}
\begin{figure*}[htbp]
  \centering
  
         \includegraphics[width=\textwidth]{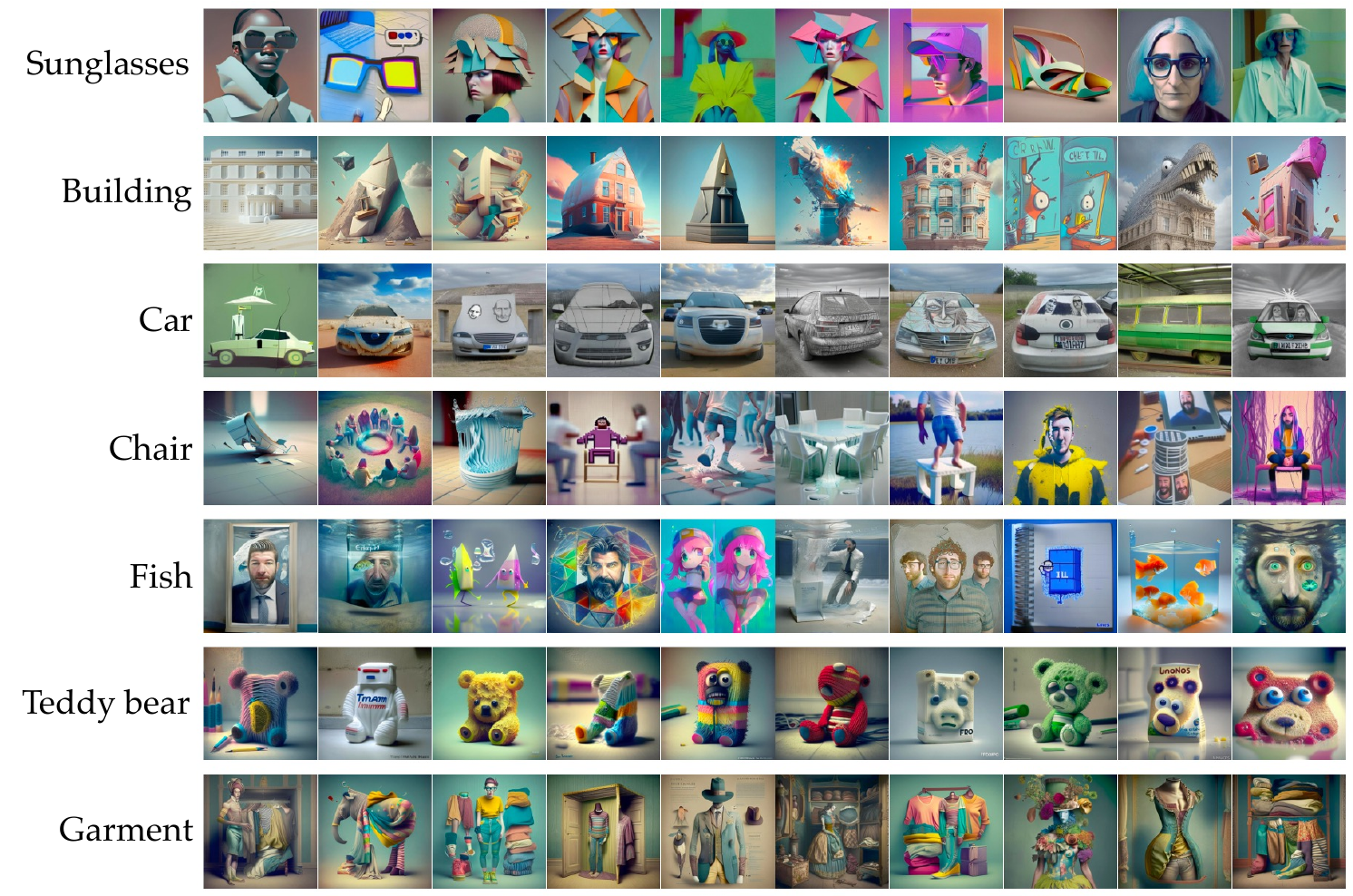}
         \caption{Uncurated samples generated from ConceptLab. Each column of samples is trained with a different initial seed.
         }
         \label{fig:app_qual_conceptlab}
  
\end{figure*}
\subsection{Quantitative Results}
We list the detailed quantitative results in Table \ref{tab:app_quant}, corresponding to each object we use in Section \ref{sec:quan}. 
FID, precision and recall are computed based on two sets of images, conventionally say, real dataset and fake dataset. For the real dataset, we use 100 images generated from SDXL with the prompt of ``a [object]". For the fake dataset, we use 100 images generated from each model and method.   
Here, we underscore again that FID and precision are interpreted as the opposite of the conventional way as our aim is to produce the object images that are distinct from the typical target object. While the image generated with our method are to be distinctive, it should also be recognized as the target object. Thus, we provide the reference scores which are computed using the fake dataset, generated from SDXL with the prompt of ``a [reference-object]" for each object. 

The overall trends are shared across the objects. For novelty metrics, our method outperforms the original generation in all objects. Especially for FID, our method does not exceed the reference score, indicating that the generated images are novel yet perceived as different objects. For `chair', our method shows low precision scores compared to the reference score, as the outlook of chairs are significantly different from the ordinary chair images. However, the high BLIP score as a usability metric defenses that the generated images with our method still look as chairs. Conversely, ConceptLab, a baseline method for comparison, presents significantly low BLIP scores for some objects as illustrated in Figure \ref{fig:qual}. This limitation of ConceptLab arises from the increased difficulty in defining sub-categories within a specific category.

Notably, our method also increases the diversity within the generated creative samples. Both LPIPS scores, which compute the distances between the generated samples in the feature space, and Vendi scores, which represent the effective number of modes among the samples, show notable improvement over the original generation across the objects. Recall scores, which are considered a measure of mode coverage within the real dataset, are comparable in most cases and show significant improvement for the Turbo model, which suffers from the mode collapse issue. 

\begin{table*}[!ht]
  \centering
  \begin{tabular}{@{}c|c|c|cc|ccc|cc@{}}
    \toprule
     \multirow{2}{*}{Object} & \multirow{2}{*}{Model} &  \multirow{2}{*}{Method} & \multicolumn{2}{c|}{Novelty}    & \multicolumn{3}{c|}{Diversity} & \multicolumn{2}{c}{Usability}    \\ \cline{4-10}

     &       &   & FID$^*$ ($\uparrow$) &  Prcs$^*$ ($\downarrow$) & Rcl ($\uparrow$) & $\text{LPIPS}$ ($\uparrow$) & Vendi ($\uparrow$) & CLIP ($\uparrow$) & BLIP ($\uparrow$)  \\
    \midrule
       \multirow{13}{*}{Chair}   &  \multirow{2}{*}{\makecell{Lightning\\(1-step)}} & Orig & 108.91& 0.81& \textbf{0.93}& 0.17& 6.58& \textbf{0.29}& \textbf{0.97} \\
   &  & Ours & \textbf{185.18}& \textbf{0.35}& 0.84& \textbf{0.27}& \textbf{8.54}& 0.27& 0.89 \\\cline{2-10}
   &  \multirow{2}{*}{Turbo} & Orig & 94.95& 0.96& 0.51& 0.20& 3.76& \textbf{0.29}& \textbf{1.00} \\
   &  & Ours & \textbf{132.72}& \textbf{0.56}& \textbf{0.62}& \textbf{0.24}& \textbf{5.92}& 0.29& 0.99 \\\cline{2-10}
   &  \multirow{2}{*}{\makecell{Lightning\\(4-step)}} & Orig & 91.55& 0.79& \textbf{1.00}& 0.20& 5.72& \textbf{0.29}& \textbf{0.99} \\
   &  & Ours & \textbf{178.05}& \textbf{0.34}& 0.76& \textbf{0.30}& \textbf{8.45}& 0.28& 0.82 \\\cline{2-10}
   &  \multirow{2}{*}{SDXL} & Orig & 104.94& 0.84& \textbf{0.97}& 0.18& 7.77& \textbf{0.29}& \textbf{0.96} \\
   &  & Ours & \textbf{158.88}& \textbf{0.60}& 0.91& \textbf{0.25}& \textbf{8.57}& 0.28& 0.87 \\\cline{2-10}
   &  Real-to-Ref  & - & 207.47& 0.87& 0.52& -& -& -& - \\\cline{2-10}
   &  ConceptLab & - & 266.58& 0.59& 0.66& 0.37& 10.46& 0.23& 0.01 \\ 
\bottomrule
\multirow{13}{*}{Teddy Bear}   &  \multirow{2}{*}{\makecell{Lightning\\(1-step)}} & Orig & 65.55& 0.99& 0.28& 0.10& 1.98& \textbf{0.29}& \textbf{1.00} \\
   &  & Ours & \textbf{82.58}& \textbf{0.80}& \textbf{0.69}& \textbf{0.23}& \textbf{2.82}& 0.28& 0.79 \\\cline{2-10}
   &  \multirow{2}{*}{Turbo} & Orig & 84.89& 0.89& 0.08& 0.14& 1.33& \textbf{0.30}& \textbf{1.00} \\
   &  & Ours & \textbf{85.11}& \textbf{0.79}& \textbf{0.53}& \textbf{0.28}& \textbf{1.71}& 0.29& \textbf{1.00} \\\cline{2-10}
   &  \multirow{2}{*}{\makecell{Lightning\\(4-step)}} & Orig & 78.07& 0.91& \textbf{0.87}& 0.20& 1.76& 0.29& \textbf{1.00} \\
   &  & Ours & \textbf{86.96}& \textbf{0.50}& 0.78& \textbf{0.30}& \textbf{3.08}& \textbf{0.29}& 0.95 \\\cline{2-10}
   &  \multirow{2}{*}{SDXL} & Orig & 67.26& 0.91& \textbf{0.98}& 0.19& 3.03& \textbf{0.29}& \textbf{1.00} \\
   &  & Ours & \textbf{97.21}& \textbf{0.84}& 0.96& \textbf{0.31}& \textbf{4.07}& 0.28& 0.89 \\\cline{2-10}
   &  Real-to-Ref  & - & 297.82& 0.71& 0.31& -& -& -& - \\\cline{2-10}
   &  ConceptLab & - & 337.86& 0.87& 0.29& 0.33& 8.12& 0.26& 0.07 \\ 
\bottomrule
\multirow{13}{*}{Garment}   &  \multirow{2}{*}{\makecell{Lightning\\(1-step)}} & Orig & 172.54& 1.00& 0.76& 0.31& 7.07& \textbf{0.27}& \textbf{1.00} \\
   &  & Ours & \textbf{193.78}& \textbf{0.93}& \textbf{0.87}& \textbf{0.39}& \textbf{8.52}& \textbf{0.27}& 0.93 \\\cline{2-10}
   &  \multirow{2}{*}{Turbo} & Orig & 212.69& 0.87& 0.15& 0.20& 5.21& \textbf{0.26}& \textbf{1.00} \\
   &  & Ours & \textbf{214.95}& \textbf{0.68}& \textbf{0.47}& \textbf{0.36}& \textbf{6.95}& 0.26& 0.93 \\\cline{2-10}
   &  \multirow{2}{*}{\makecell{Lightning\\(4-step)}} & Orig & 165.04& 0.91& \textbf{0.93}& 0.29& 7.58& \textbf{0.26}& \textbf{0.98} \\
   &  & Ours & \textbf{176.02}& \textbf{0.72}& 0.92& \textbf{0.37}& \textbf{8.78}& 0.26& 0.89 \\\cline{2-10}
   &  \multirow{2}{*}{SDXL} & Orig & 167.05& 0.89& \textbf{0.95}& 0.22& 8.31& \textbf{0.27}& \textbf{0.94} \\
   &  & Ours & \textbf{196.13}& \textbf{0.74}& 0.94& \textbf{0.38}& \textbf{9.10}& 0.26& 0.81 \\\cline{2-10}
   &  Real-to-Ref  & - & 232.91& 0.83& 0.80& -& -& -& - \\\cline{2-10}
   &  ConceptLab & - & 225.85& 0.89& 0.71& 0.37& 7.79& 0.26& 0.66 \\ 
\bottomrule
\multirow{13}{*}{Car}   &  \multirow{2}{*}{\makecell{Lightning\\(1-step)}} & Orig & 92.83& 0.84& 0.90& 0.36& 4.78& 0.25& \textbf{0.89} \\
   &  & Ours & \textbf{111.43}& \textbf{0.61}& \textbf{0.93}& \textbf{0.42}& \textbf{5.25}& \textbf{0.26}& 0.88 \\\cline{2-10}
   &  \multirow{2}{*}{Turbo} & Orig & 131.62& 0.77& 0.45& 0.30& 3.08& 0.26& \textbf{1.00} \\
   &  & Ours & \textbf{150.14}& \textbf{0.22}& \textbf{0.96}& \textbf{0.46}& \textbf{4.89}& \textbf{0.26}& 0.84 \\\cline{2-10}
   &  \multirow{2}{*}{\makecell{Lightning\\(4-step)}} & Orig & 86.82& 0.85& 0.89& 0.39& 3.90& 0.25& \textbf{1.00} \\
   &  & Ours & \textbf{110.63}& \textbf{0.55}& \textbf{0.95}& \textbf{0.43}& \textbf{4.70}& \textbf{0.26}& 0.95 \\\cline{2-10}
   &  \multirow{2}{*}{SDXL} & Orig & 119.38& 0.54& 0.96& 0.30& 5.47& 0.26& \textbf{0.88} \\
   &  & Ours & \textbf{137.24}& \textbf{0.39}& \textbf{0.98}& \textbf{0.34}& \textbf{6.24}& \textbf{0.26}& 0.77 \\\cline{2-10}
   &  Real-to-Ref  & - & 220.79& 0.31& 0.38& -& -& -& - \\\cline{2-10}
   &  ConceptLab & - & 151.87& 0.47& 0.79& 0.40& 6.70& 0.27& 0.71 \\ 
\bottomrule
\multirow{13}{*}{Building}   &  \multirow{2}{*}{\makecell{Lightning\\(1-step)}} & Orig & 179.90& 0.83& 0.56& 0.34& \textbf{6.12}& \textbf{0.26}& \textbf{1.00} \\
   &  & Ours & \textbf{242.10}& \textbf{0.57}& \textbf{0.64}& \textbf{0.37}& 6.07& 0.24& 0.97 \\\cline{2-10}
   &  \multirow{2}{*}{Turbo} & Orig & 208.01& 0.86& 0.17& 0.28& 4.31& \textbf{0.24}& \textbf{1.00} \\
   &  & Ours & \textbf{237.44}& \textbf{0.32}& \textbf{0.82}& \textbf{0.48}& \textbf{5.19}& \textbf{0.24}& \textbf{1.00} \\\cline{2-10}
   &  \multirow{2}{*}{\makecell{Lightning\\(4-step)}} & Orig & 165.12& 0.84& \textbf{0.89}& 0.35& 6.17& \textbf{0.25}& 0.98 \\
   &  & Ours & \textbf{222.89}& \textbf{0.65}& 0.75& \textbf{0.37}& \textbf{6.37}& 0.24& \textbf{0.99} \\\cline{2-10}
   &  \multirow{2}{*}{SDXL} & Orig & 182.38& 0.76& \textbf{0.96}& 0.30& 8.00& \textbf{0.24}& \textbf{0.97} \\
   &  & Ours & \textbf{200.18}& \textbf{0.74}& 0.87& \textbf{0.32}& \textbf{8.64}& 0.24& \textbf{0.97} \\\cline{2-10}
   &  Real-to-Ref  & - & 281.64& 0.09& 0.80& -& -& -& - \\\cline{2-10}
   &  ConceptLab & - & 274.19& 0.43& 0.77& 0.37& 8.98& 0.23& 0.13 \\ 
\bottomrule

  \end{tabular}
  \caption{Object-wise quantitative results. `Prcs' and `Rcl' refer to precision and recall, respectively.  }
  \label{tab:app_quant}
\end{table*}

\clearpage
\subsection{User Study}

\begin{figure*}[!ht]
  \centering
  
         \includegraphics[width=0.8\textwidth]{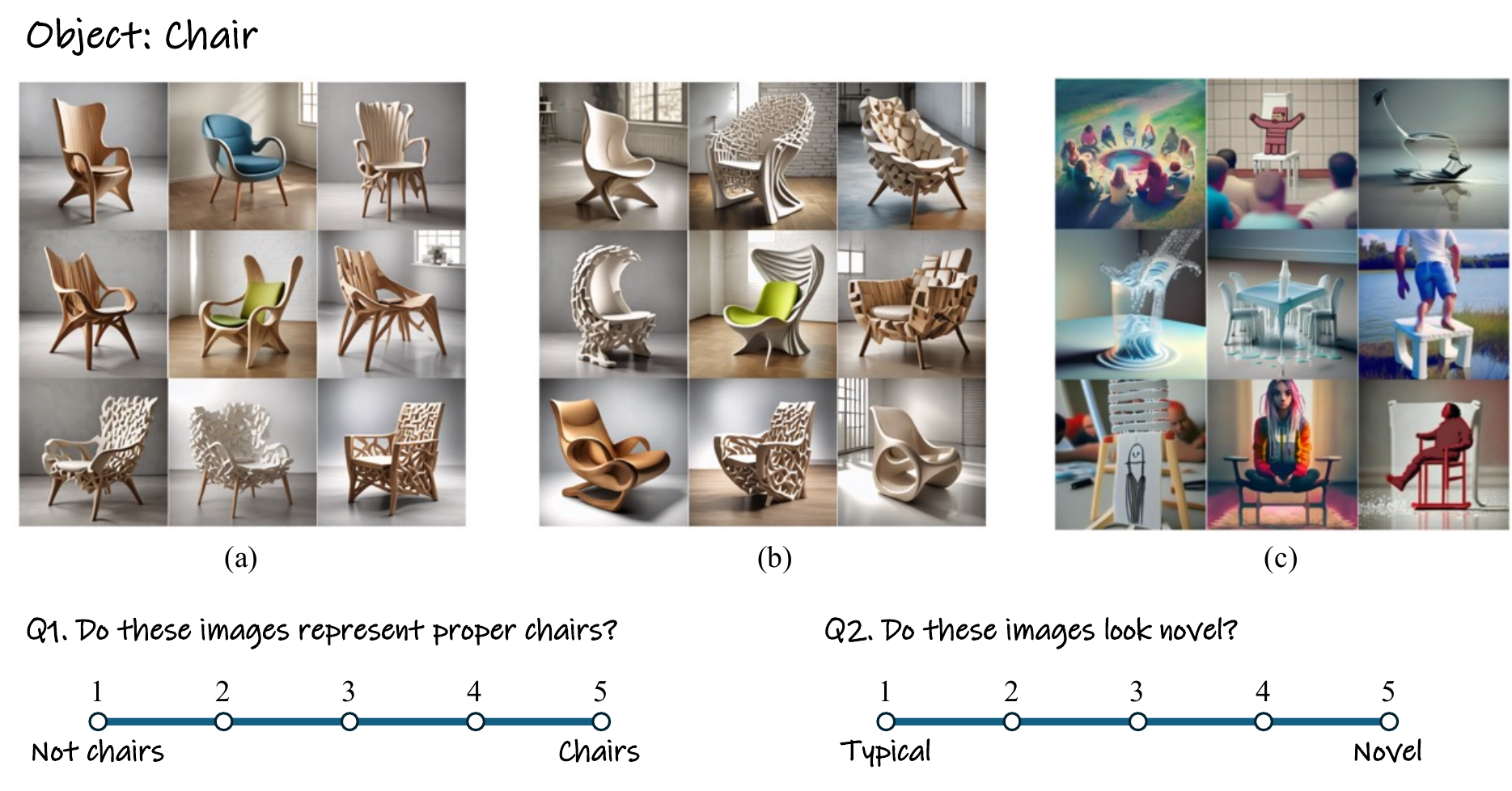}
         \caption{ Example of user study questionnaires for the object `Chair'. The questions are posed repeatedly for each set of images. (a) SDXL-Turbo (Original). (b) SDXL-Turbo (Ours). (c) ConceptLab. Images are carefully curated as the best to represent each method.  
         }
         \label{fig:app_user_study_sample}
\end{figure*}
\vskip -0.5cm
\begin{figure}[!ht]
  \centering
  
         \includegraphics[width=0.9\textwidth]{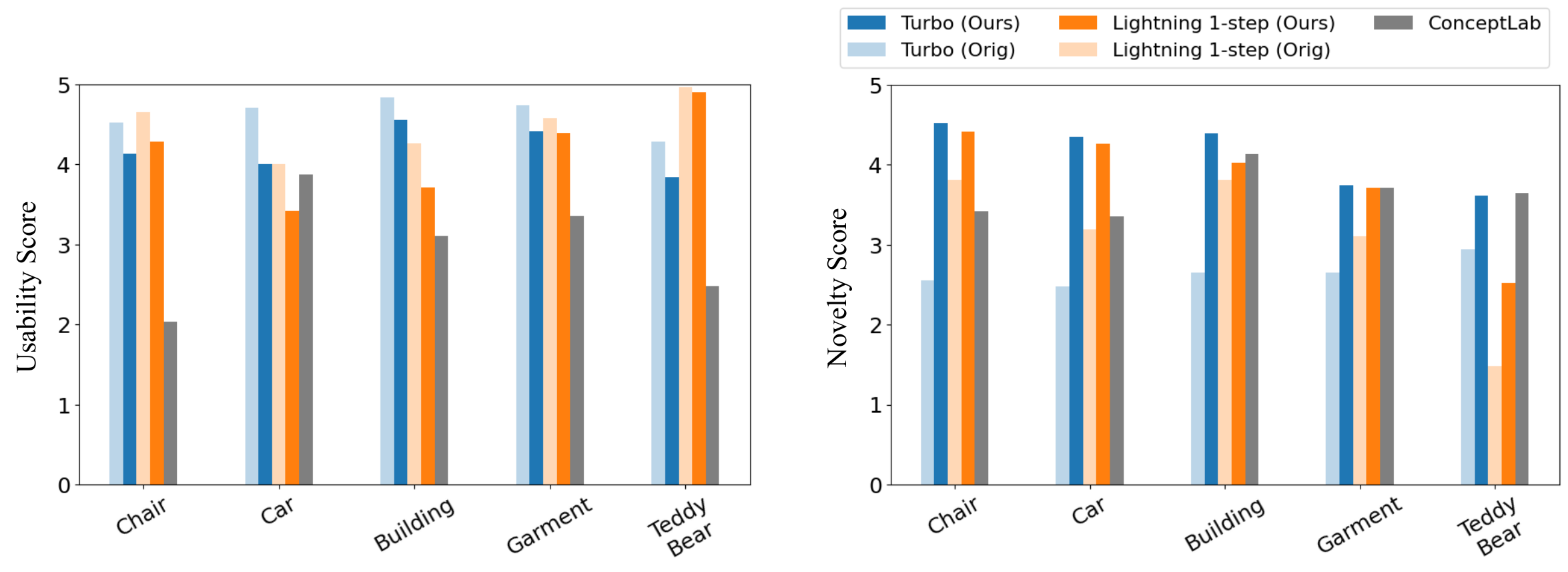}
         \caption{Object-wise results of user study. Our method improves the original results in terms of both usability and novelty for all objects. 
         }
         \label{fig:app_user_study}
  
\end{figure}
We conduct a user study to evaluate the creativity of generated samples for each method with human perception. Given a set of images, we ask participants questions regarding two main aspects of creativity: usability and novelty. As an example illustrated in Figure \ref{fig:app_user_study_sample}, with a target object `chair', we ask (1) whether these images represent proper chairs and (2) whether these images look novel. The responses are collected using a 5-level Likert scale.  
For a target object, images are generated from the model's default setting, marked as `Orig', and from \textbf{C3}, marked as `Ours'. Additionally, images are generated from ConceptLab for baseline comparison. For each image set, we carefully select 9 images that appear to be the most creative within the generated images for each method. While each method is anonymized in the questions, we denote each method in the result as `Turbo (Orig/Ours)', `Lightning (1-step) (Orig/Ours)', and `ConceptLab', respectively.


In total, 31 participants have responded. 
We summarize the responses for each object in Figure \ref{fig:app_user_study}. The blue bar plots represent results corresponding to Turbo, while the yellow bar plots represent results corresponding to Lightning (1-step). The darker color represents \textbf{C3} marked as `Ours', while the lighter color represents the default generation marked as `Orig'. The results corresponding to ConceptLab are presented with the gray bar plots. Our method significantly enhances novelty in all cases, with only a relatively small reduction in usability scores. Notably, our method outperforms ConceptLab in usability scores and achieves higher or comparable novelty scores. 
\clearpage

\section{Types of Creativity}
\begin{figure*}[!ht]
  \centering
    \vskip -0.5cm
         \includegraphics[width=0.75\textwidth]{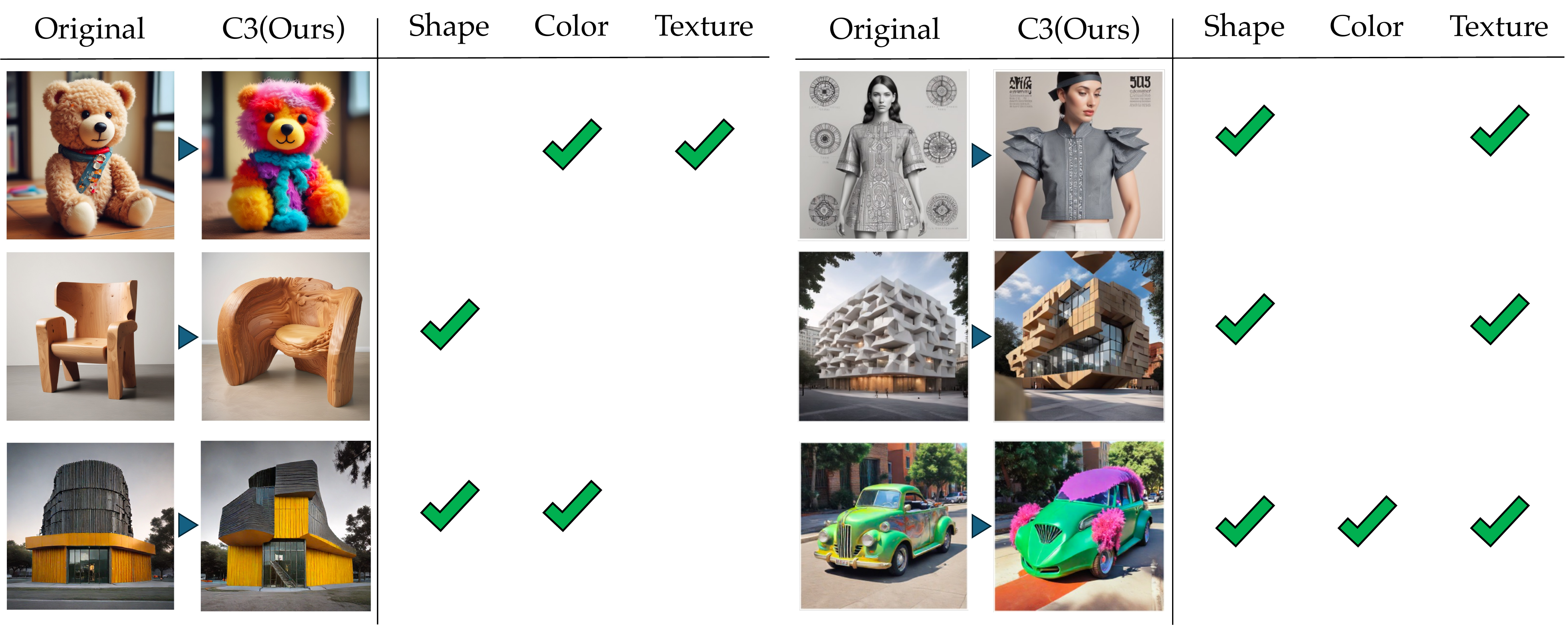}
         \caption{Types of creativity classified by GPT 4o of samples generated by the proposed method compared to samples generated by the original models. Responses are multiple-choice, among ‘Shape’, ‘Texture’, and ‘Color’. 
         }
         \label{fig:app_type_1}
  
\end{figure*}

Here, we present the categories of creativity amplified in samples generated using the proposed method. All images are generated using the prompt ``a creative [object]." The teddy bear, garment, and chair images are generated based on the backbone model Lightning (4-step), while a building image (left) and the car images are generated using the Lightning (1-step) model. Another building image (right) is generated with the Turbo.

Same as the settings of Section~\ref{sec:exp_creativity_type}, we utilize GPT4o~\cite{achiam2023gpt} to obtain responses. The exact question posed is:

\textit{``Please identify the components that contribute to the creativity of the second image (ours) compared to the first image (original). The components can be selected from shapes, colors, and textures. If none apply, state no.”}

As illustrated in Figure~\ref{fig:app_type_1}, images generated with \textbf{C3} demonstrate enhancements in various aspects of creativity. For instance, in the case of ``a creative teddy bear," the generated image becomes more creative through enriched color (a more vibrant, colorful body and scarf) and texture (a fluffy appearance).

\section{Failure Cases}
\label{sec:app_failure}
\begin{figure*}[!ht]
  \centering
  \vskip -0.5cm
  
         \includegraphics[width=0.75\textwidth]{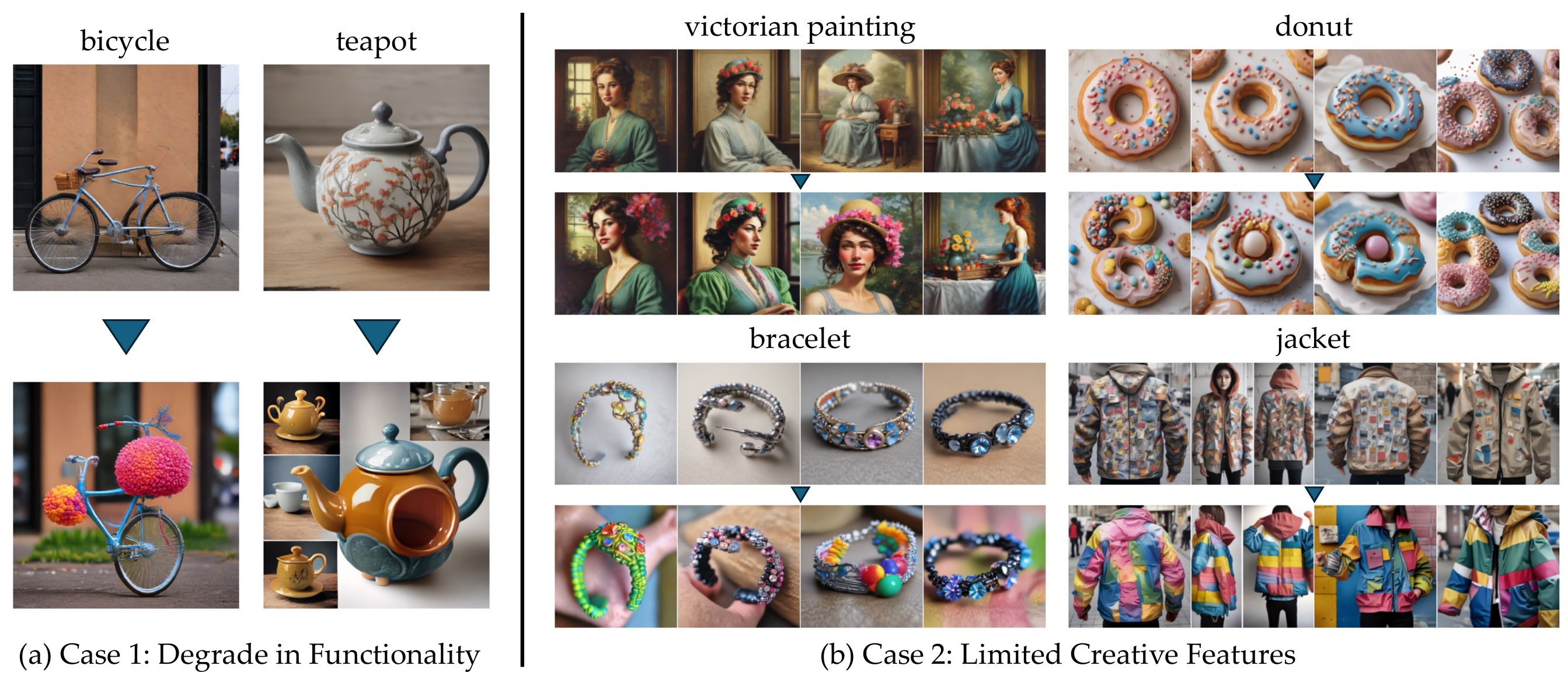}
         \caption{Failure cases of \textbf{C3}. Lightning (1-step) is used for `bicycle', `victorian painting', and `bracelet' and Turbo is used for `teapot', `donut' and `jacket'.
         }
         \label{fig:app_failure}
  
\end{figure*}
While we show \textbf{C3} successfully enhances the creative generations of pretrained Stable Diffsuion-based models, there exist failure cases. We present two main failure cases in Figure \ref{fig:app_failure}. In Case 1, \textbf{C3} produces genuinely novel images, but this comes at the cost of reduced functionality. This occurs primarily due to the limitations of the current usability score, which is inadequate for assessing detailed functionality. In Case 2, the generated images exhibit enhanced creativity (e.g., modern fashion in Victorian-style painting), but they do not significantly deviate from common patterns: a woman in a painting, a round-shape donut, a colorful bracelet, and a color-patched vinyl jacket. We presume that these patterns originate from biases present in the pre-trained models. We leave the generation of creative outputs that mitigate model bias for future work.

\section{Extension on Non-SD Models}
\label{sec:app_nonsd}
\begin{figure}[h!]
  \centering
         \includegraphics[width=0.8\columnwidth]{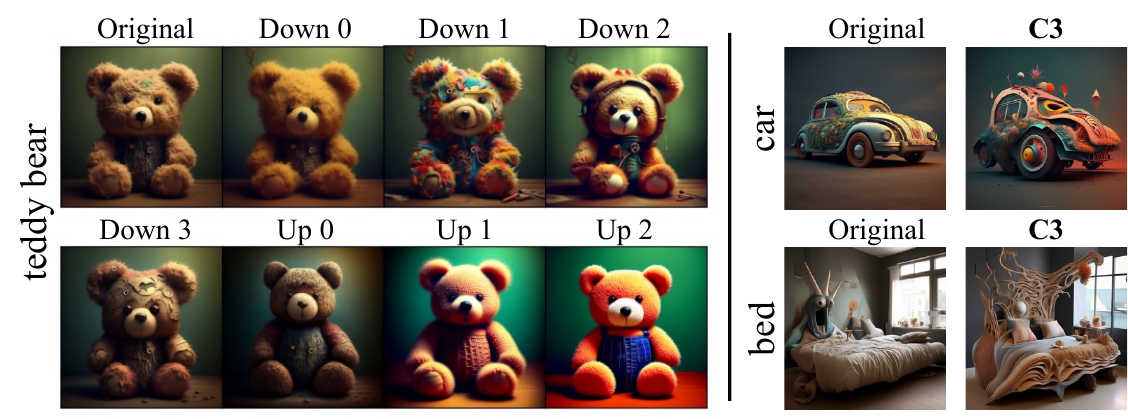}
         \caption{(Left) Block-wise amplified images on Kandinsky 3.0. (Right) Results of C3 applied on Down blocks of Kandinsky 3.0.}
         \label{fig:kandinsky}  
\end{figure}

\noindent\textbf{Unet-based Model.} In Figure~\ref{fig:kandinsky}, we applied C3 to Kandinsky 3.0. Similarly to Stable Diffusion XL (SDXL), structural/color variation occurs more on down blocks, while up blocks increase the filter effect, such as contrast.  

\begin{figure}[h!]
  \centering
         \includegraphics[width=0.8\columnwidth]{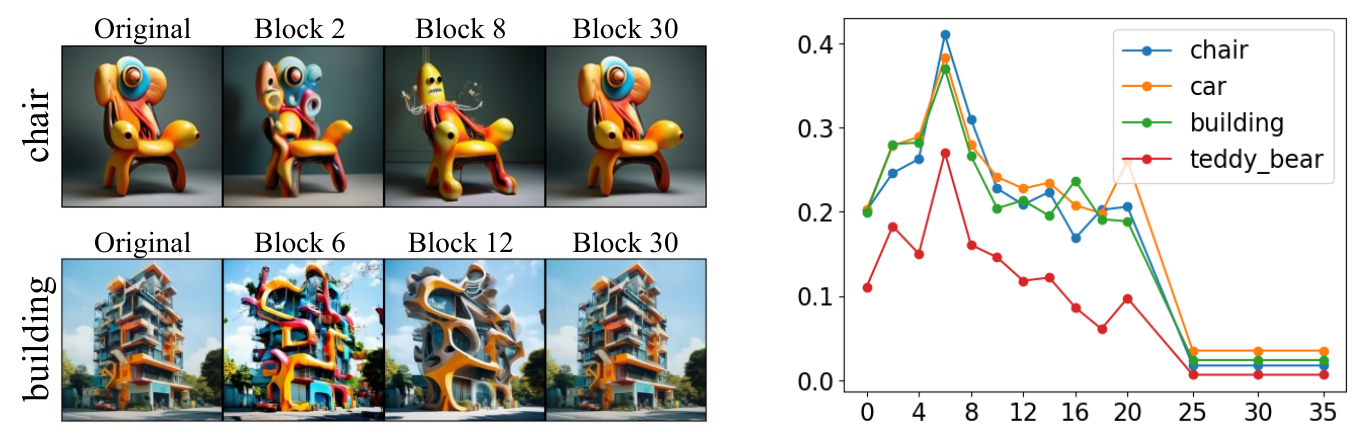}
         \caption{(Left) Images generated with amplified features for the ith transformer block of Hunyuan-DiT. (Right) Block-wise LPIPS score between the original image and the amplified image.}
         \label{fig:dit}  
\end{figure}

\noindent\textbf{Transformer-based Model.}
In Figure~\ref{fig:dit}, we applied C3 on Hunyuan-DiT.
Amplifying blocks 5–20 yields the most variation, while blocks beyond 20 show minimal change even at maximum amplification. Nevertheless, the impact of C3 on transformer-based models requires careful study, which we leave for future work.

\section{Related Work}
\label{sec:app_related}
\subsection{Stable Diffusion Models}
Diffusion models \cite{ho2020denoising,songdenoising} learn to generate images from random noise through a denoising process, demonstrating stable training and remarkable performance in image and video generation compared to GANs. Latent Diffusion model \cite{rombach2022high}, instead of directly processing images during the denoising process, learns the encoded latent vectors of the images, successfully reducing the size of the model. Stable Diffusion XL (SDXL) \cite{podell2023sdxl}, a widely used Latent Diffusion model, is publicly available with accessible source code and trained models.
Diffusion models undergo denoising processes in T steps in order to generate a sample. To accelerate high-quality sample generation, distilled models have been developed. SDXL-Turbo (Turbo) \cite{sauer2025adversarial} employs Adversarial Diffusion Distillation (ADD) to condense the multi-step denoising process of a large pre-trained teacher model into 1-4 steps while maintaining high quality. However, due to the limitations of adversarial training, Turbo cannot prevent mode collapse. SDXL-Lightning (Lightning) \cite{lin2024sdxl} combines ADD with progressive distillation to address mode collapse while quickly generating high-quality samples in one or a few steps.

SDXL and its distilled variants share a U-net structure in the backbone to generate a latent noise. The U-net structure is composed of three down blocks, decreasing the resolution of the internal feature maps while increasing the number of channels, a middle block, and three up blocks, increasing the resolution of the feature maps again.

Recent research~\cite{jeong2024visual} indicates that up blocks in Stable Diffusion models are primarily associated with style while the structure is preserved. This aligns with our observation that up blocks minimally alter the creative style when amplified, although some filter effects are introduced. Other models may exhibit different block characteristics; for instance, Disco-diff~\cite{xu2024disco}, influenced by StyleGAN, employs discrete latents for each block and trains end-to-end, potentially redefining block roles.

\subsection{Creative Generation}
Research on achieving creative generations in generative models has been continuously advancing. Based on GANs, creative generations are encouraged by employing contrastive loss or diversity loss from existing categories or samples ~\cite{nobari2021creativegan,sbai2018design,elgammal2017can}. 
Recent advances in generative modeling have aimed to balance creativity with diversity in image generation, focusing on approaches that allow inspiration from existing concepts without direct replication. ProCreate~\cite{lu2024procreate}, an energy-based approach, proposes guiding diffusion model outputs away from reference images in the latent space, thus improving diversity and concept fidelity in few-shot settings. This method prevents training data replication and has been shown to enhance sample creativity across various artistic styles and categories. On the other hand, Inspiration Tree~\cite{vinker2023concept} introduces a structured decomposition of concepts, where a hierarchical tree structure captures different visual aspects of a given concept. 
Adding to this line of creative generative techniques, ConceptLab~\cite{richardson2023conceptlab} leverages a Vision-Language Model (VLM) with diffusion priors to further push the boundaries of novel concept generation within broad categories. By iteratively applying constraints that differentiate generated concepts from existing category members, ConceptLab enhances the creation of unique, never-before-seen concepts, enabling hybridization and exploration within a given category. 
While these approaches represent significant advancements in generating creatively inspired outputs, they necessitate burdensome additional training or optimization. To the best of our knowledge, there is no training-free approach for generating creative samples.

\subsection{Feature Map Manipulation}
GAN Dissection~\cite{bau2019gan} pioneered techniques to visualize and control the inner workings of GANs by identifying ``interpretable units” that correspond to distinct objects within generated images. This approach enables precise feature control, allowing specific objects to be added or removed within scenes, making it effective for targeted scene composition. 
Expanding on internal feature manipulation to the text-to-image diffusion models, research into internal features of diffusion models is advancing rapidly~\cite{voynov2023p+,kwon2022diffusion,hertz2022prompt,cao2023masactrl}. Especially, P+~\cite{voynov2023p+} takes a step further by introducing multi-layered conditioning, enabling flexible visual manipulation and enhanced image customization through layer-specific control. Our work builds on these foundations by exploring feature manipulation in the Fourier domain for even greater creativity control. 
FreeU~\cite{si2024freeu} also operates within the Fourier domain, leveraging Fourier transforms on skip connections to reduce low-frequency information, ultimately improving image fidelity in diffusion models. In contrast, our approach amplifies creativity by applying Fourier-based manipulation directly to the backbone features in the specific blocks rather than skip connections. This distinction allows our method to focus on enhancing creative aspects, making it suited for generating novel and expressive images.



\end{document}